\documentclass[sigconf,10pt, nonacm, screen]{acmart}
\PassOptionsToPackage{cache=false}{minted}

\usepackage[most]{tcolorbox}
\tcbuselibrary{minted,skins,breakable}

\usepackage{amsmath}
\usepackage{paralist}
\definecolor{set2cyan}{RGB}{114, 182, 161}
\definecolor{set2orange}{RGB}{232, 149, 117}
\definecolor{set2blue}{RGB}{149, 163, 195}
\definecolor{set2purple}{RGB}{219, 149, 192}
\definecolor{set2green}{RGB}{162, 199, 101}
\definecolor{set2yellow}{RGB}{255, 217, 47}
\definecolor{set2brown}{RGB}{229, 196, 148}
\definecolor{set2gray}{RGB}{88, 88, 88}
\definecolor{set2peach}{RGB}{253, 192, 134}
\definecolor{set2green2}{RGB}{127, 201, 127}
\definecolor{tealblue}{RGB}{54, 144, 192}
\definecolor{softcoral}{RGB}{242, 133, 108}
\definecolor{amethystpurple}{RGB}{153, 102, 204}

\definecolor{darkgreen}{RGB}{0, 70, 0}
\definecolor{darkblue}{RGB}{20, 33, 61}
\definecolor{darkred}{rgb}{0.65,0,0}
\definecolor{mygreen}{RGB}{208, 244, 222}
\definecolor{myp}{RGB}{253, 226, 228}
\definecolor{myb}{RGB}{198, 222, 241}
\definecolor{mymagenta}{RGB}{255,0,255} 
\definecolor{myo}{RGB}{255,200,0}
\definecolor{pastelblue}{RGB}{102,153,204}
\definecolor{forestgreen}{rgb}{0.13, 0.65, 0.13}

\usepackage[capitalise, nameinlink]{cleveref}
\usepackage{xcolor}
\usepackage{graphicx}
\usepackage{algorithm}
\usepackage{algorithmicx}
\usepackage[noend]{algpseudocode} 
\crefname{algorithm}{Alg.}{Alg.}
\Crefname{algorithm}{Alg.}{Alg.}
\crefname{section}{\S\!}{\S\S\!}
\Crefname{section}{Sec.}{Sec.}
\crefname{figure}{Fig.}{Fig.}
\Crefname{figures}{Fig.}{Fig.}
\crefname{equation}{Eqn.}{Eqn.}
\Crefname{equation}{Eqn.}{Eqn.}
\crefname{defn}{definition}{definitions}

\newcommand{\eg}{\emph{e.g.}\xspace}

\newcommand{\ETAL}{\emph{et al.}\xspace}

\newif\ifremovecomments
\ifremovecomments
    \usepackage[disable, textsize=small]{todonotes}
\else
    \usepackage[textsize=small]{todonotes}
\usepackage{xspace}
\fi
\newcommand{\sysname}{\textsc{Robusta}\xspace}
\usepackage{enumitem}
\usepackage{varwidth}

\usepackage{dblfloatfix}
\definecolor{darkgreen}{RGB}{20, 33, 61}
\definecolor{darkblue}{RGB}{20, 33, 61}
\definecolor{darkred}{rgb}{0.65,0,0}

\lstdefinestyle{CStyle}{
    language=C,
    basicstyle=\ttfamily\small,
    commentstyle=\color{green!40!black},
    keywordstyle=\bfseries\color{black},
    numbers=none,
    numberstyle=\tiny,
    numbersep=5pt,
    frame=single,
    showstringspaces=false,
    numberstyle=\scriptsize\color{gray},
    xleftmargin=0pt,
    xleftmargin=.75em,
    framexleftmargin=.5em,
    upquote=true,
    showspaces=false,
    showtabs=false,
    tabsize=2,
    columns=flexible,
    frame=none,
    literate={€}{{}}1,
    morekeywords={printf, scanf, int, void, if, else, while, for, return, bool}
}
\lstdefinestyle{PythonStyle}{
    language=Python,
    basicstyle=\ttfamily,
    commentstyle=\color{green!40!black},
    keywordstyle=\bfseries\color{blue},
    stringstyle=\color{darkred},
    numbers=none,
    numberstyle=\tiny,
    numbersep=5pt,
    frame=single,
    showstringspaces=false,
    numberstyle=\scriptsize\color{gray},
    xleftmargin=0pt,
    xleftmargin=.75em,
    framexleftmargin=.5em,
    upquote=true,
    showspaces=false,
    showtabs=false,
    tabsize=2,
    columns=flexible,
    frame=none,
    literate={€}{{}}1
}
\lstdefinestyle{TextStyle}{
    language={},
    basicstyle=\ttfamily\small,
    keywordstyle=\color{black},
    numbers=none,
    numberstyle=\tiny,
    numbersep=5pt,
    frame=single,
    showstringspaces=false,
    numberstyle=\scriptsize\color{gray},
    xleftmargin=0pt,
    xleftmargin=.75em,
    framexleftmargin=.5em,
    upquote=true,
    showspaces=false,
    showtabs=false,
    tabsize=2,
    columns=flexible,
    frame=none,
    literate={€}{{}}1
}

\title{Robust Heuristic Algorithm Design with LLMs}

\author{Pantea Karimi}
\affiliation{\institution{MIT}\city{}\country{}}

\author{Dany Rouhana}
\affiliation{\institution{Microsoft}\city{}\country{}}

\author{Pooria Namyar}
\affiliation{\institution{University of Southern California}\city{}\country{}}

\author{\hspace{-0.07cm}Siva Kesava Reddy Kakarla}
\affiliation{\institution{Microsoft Research}\city{}\country{}}

\author{Venkat  Arun}
\affiliation{\institution{The University of Texas at Austin}\city{}\country{}}

\author{Behnaz Arzani}
\affiliation{\institution{Microsoft Research}\city{}\country{}}

\usepackage{pgfplots}

\usepackage{pgfplotstable}
\usetikzlibrary{fit,backgrounds,decorations.pathreplacing}
\usepgfplotslibrary{groupplots,statistics,fillbetween}
\pgfplotsset{compat=1.18}

\usepackage{tikz}
\usetikzlibrary{
  shapes.misc, shapes.geometric, shapes.multipart, shapes.arrows,
  positioning, quotes, arrows.meta, decorations.pathmorphing, 
  matrix, graphs, fit, calc, automata, shadows
}

\usepackage{graphicx}
\DeclareGraphicsExtensions{.pdf,.png,.jpg}

\begin{document}

\renewcommand{\shortauthors}{Karimi \textit{et al.}}
\acmConference{}
\acmISBN{}
\maketitle

\noindent \textbf{Abstract ---} We posit that we can generate more robust and performant heuristics if we augment approaches using LLMs for heuristic design with tools that explain why heuristics underperform and suggestions about how to fix them. We find even simple ideas that (1) expose the LLM to instances where the heuristic underperforms; (2) explain why they occur; and (3) specialize design to regions in the input space, can produce more robust algorithms compared to existing techniques~---~the heuristics we produce have a $\sim28\times$ better worst-case performance compared to FunSearch, improve average performance, and maintain the runtime.

\section{Introduction}

This paper asks whether LLMs can help design~\emph{robust} heuristic (approximate) algorithms and whether ``old-school'' modeling techniques, like heuristic analysis and combinatorial reasoning, can help them do so more effectively\footnote{We discuss why LLM-based approaches are the right technique in~\S\ref{sec:future}}. 

We need to ``robustify'' heuristics. Deployed heuristics can fail in certain important edge-cases, which can cause catastrophic impact~\cite{arun2022starvation}, but operators often deploy them anyway~\cite{metaopt, soroush, danna, swan, cassini, taccl, minaQueue, islip, themis, liangyu} because they are faster or more efficient than the optimal. Prior work has found instances of such~\emph{deployed} heuristics that have severe performance problems under practical workloads~\cite{arun2022starvation, metaopt, raha}.

 It is hard to design robust heuristics. Researchers have tried to improve certain heuristics~\cite{swan, soroush, ncflow, pop, danna, islip,aifo,sp-pifo,packs, ishai,bennell2013genetic,falkenauer1992genetic,kroger1995guillotineable, themis, cassini}. 
 But the heuristic's performance is tightly coupled with the workloads and hardware~---~operators have to often re-design or change heuristics as those parameters change. We need to reason across the problem structure, the workload, the hardware, and the behavior of other systems that interact with the heuristic to robustify it.
 
Our goal is to automate this process so that operators can easily create robust heuristics for any hardware or workload. This allows us to lower the risk of deploying these heuristics.

Recent works use LLMs to improve heuristics~\cite{liu2024evolution, gao2024alpha, funsearch}, and companies have deployed the ``synthetic'' algorithms they produce~\cite{gao2024alpha}. These solutions use LLMs in a search process (\eg, genetic search) where the LLM produces new heuristics based on feedback on the performance of those it produced so far (\cref{sec:background}). These tools ``evaluate'' the code the LLM produces on random samples from the input space (or samples from their production workloads)~---~their goal is to improve the average performance of the heuristic and they often ignore important corner cases.
We find these solutions only scratch the surface of what's possible: even simple ideas that strategically select inputs (to evaluate the code on) improve the worst-case performance of the heuristics they generate in each step by $\ge 20\%$ (\cref{fig:oneshot})\footnote{We find the heuristics we generate (surprisingly) do not harm the runtime and even improve the average performance of the heuristic (\autoref{table:avg-perf}).}

\tikzset{
  shaded/.style={
    draw,
    minimum width=1cm,
    minimum height=.7cm,
    rounded corners,
    fill=gray!20,
   align=center,
    semithick,
    text=black,
    drop shadow={
      shadow xshift=0.5mm,
      shadow yshift=-0.5mm,
      opacity=0.5
    }
  }
}

\tikzset{
  plain/.style={
    draw,
    minimum width=2cm,
    minimum height=.9cm,
    rounded corners,
    align=center,
    fill=teal!5,
    semithick,
    text=black,
    drop shadow={
      shadow xshift=0.5mm,
      shadow yshift=-0.5mm,
      opacity=0.5
    }
  }
}
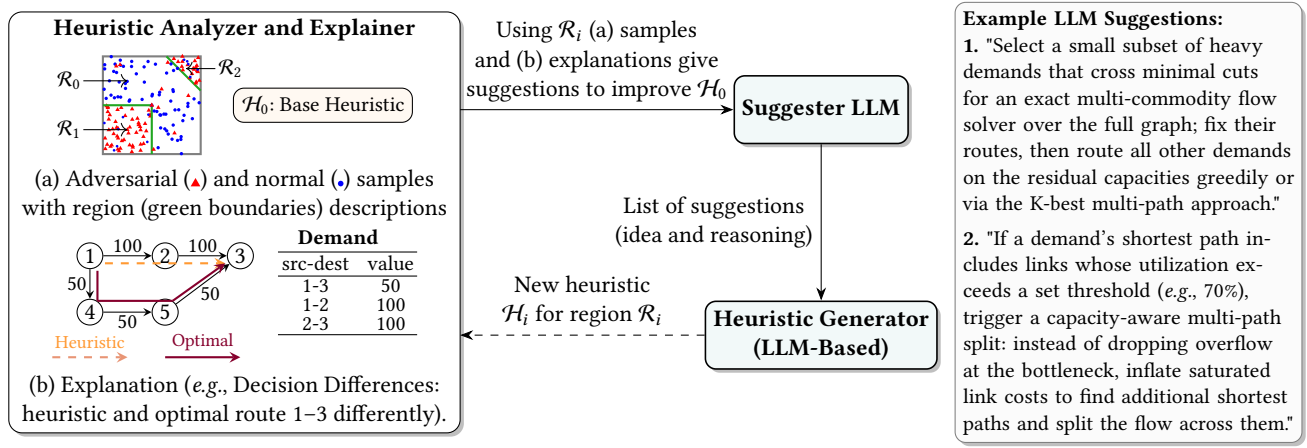
\begin{figure*}[!t]
  \centering
  \small
  \begin{tikzpicture}[every arrow/.style={}, scale=0.88]
    \begin{scope}[xshift=-1cm]
      \node[inner sep=3pt,fill=white!7] (background) at (0,0) {
            \begin{tikzpicture}[every arrow/.style={}, scale = 0.8]
    \node[inner sep=3pt,fill=white!7] (background) at (0,0) {
        \begin{tikzpicture}
            \draw[rounded corners=5pt, fill=white, drop shadow={shadow xshift=0.5mm, shadow yshift=-0.5mm, opacity=0.5 }] (-3,-2.8) rectangle (3,2.8);
            \node[inner sep=5pt] (analyzer) at (0,0) {
                \begin{tikzpicture}
                  \node at (5,3.5) {\textbf{Heuristic Analyzer and Explainer}};
                  
                  \node[inner sep=0pt, label={[align=center]below:{(a) Adversarial (\tikz{\node[red, fill=red, regular polygon, regular polygon sides=3, inner sep=0.75pt] {};}) and normal (\tikz{\node[blue, fill=blue, circle, inner sep=0.75pt] {};}) samples \\with region (green boundaries) descriptions}}]  (node1) at (5,2.5) { 
                    \begin{tikzpicture}[scale=0.65
                    ]
                    \draw[gray, thick] (0,0) rectangle (2,2);
                    \draw[forestgreen, thick] (1.3,2) -- (2,1.3);
                    \foreach \x/\y in {
                        0.375/0.951,
                        0.732/0.599,
                        0.156/0.156,
                        0.1/0.866,
                        0.601/0.708,
                        0.1/0.970,
                        0.832/0.212,
                        0.182/0.183,
                        0.304/0.525,
                        0.432/0.291,
                        0.612/0.139,
                        0.292/0.366,
                        0.456/0.785,
                        0.200/0.514,
                        0.592/0.046,
                        0.608/0.171,
                        0.1/0.949,
                        0.966/0.808,
                        0.305/0.098,
                        0.684/0.440,
                        0.122/0.495,
                        0.3/0.909,
                        0.259/0.663,
                        0.312/0.520,
                        0.547/0.185,
                        0.970/0.775,
                        0.939/0.895,
                        0.598/0.922,
                        0.088/0.196,
                        0.045/0.325,
                        0.389/0.271,
                        0.829/0.357,
                        0.281/0.543,
                        0.141/0.802,
                        0.075/0.987,
                        0.772/0.199,
                        0.22/0.415,
                        0.5/0.5,
                        0.55/0.55,
                        0.707/0.729,
                        0.771/0.074,
                        0.358/0.116,
                        0.863/0.623,
                        0.331/0.064,
                        0.311/0.325,
                        0.730/0.638,
                        0.887/0.472,
                        0.120/0.713,
                        0.761/0.561,
                        0.771/0.494,
                        0.523/0.428,
                        0.05/0.108
                    } {
                        \node[red, fill=red, regular polygon, regular polygon sides=3, inner sep=0.35pt, minimum size=1pt] at (\x,\y) {};
                    }
                    \foreach \x/\y in {
                         0.031/0.636,
                            0.314/0.509,
                            0.908/0.249,
                            0.410/0.756,
                            0.229/0.077,
                            0.290/0.161,
                            0.930/0.808
                    } {
                        \node[blue, fill=blue, circle, inner sep=0.35pt, minimum size=1pt] at (\x,\y) {};
                    }
                    \draw[forestgreen, thick] (0,1) -- (1,1);
                    \draw[forestgreen, thick] (1,0) -- (1,1);
                    \foreach \x/\y in {
                        1.615/1.792,
                        1.636/1.721,
                        1.849/1.755,
                        1.774/1.560,
                        1.881/1.908,
                        1.927/1.706,
                        1.734/1.826,
                        1.452/1.952,
                        1.910/1.476,
                        1.922/1.811,
                        1.626/1.894,
                        1.972/1.507,
                        1.901/1.781,
                        1.949/1.972,
                        1.822/1.645,
                        1.900/1.451,
                        1.865/1.732,
                        1.940/1.684,
                        1.626/1.999,
                        1.538/1.890,
                    } {
                        \node[red, fill=red, regular polygon, regular polygon sides=3, inner sep=0.35pt, minimum size=1pt] at (\x,\y) {};
                    }
                    \foreach \x/\y in {
                            1.725/1.899,
                            1.747/1.842,
                            1.939/1.731,
                            1.661/1.930,
                            1.788/1.577
                    } {
                        \node[blue, fill=blue, circle, inner sep=0.35pt, minimum size=1pt] at (\x,\y) {};
                    }
                    \foreach \x/\y in {
                        0.497/1.488,
                        0.067/1.140,
                        1.525/1.754,
                        0.684/1.643,
                        0.221/1.693,
                        1.595/0.300,
                        0.459/1.445,
                        1.440/1.282,
                        1.388/1.085,
                        0.363/1.817,
                        1.167/0.802,
                        0.924/1.895,
                        0.307/1.172,
                        1.012/1.223,
                        0.036/1.744,
                        1.864/1.130,
                        1.393/1.845,
                        1.414/0.305,
                        1.153/1.213,
                        0.848/1.473,
                        0.249/1.842,
                        1.740/1.038,
                        1.183/0.798,
                        1.606/0.009,
                        1.075/1.840,
                        1.475/0.904,
                        1.830/0.725,
                        1.161/1.265,
                        0.026/1.327,
                        0.356/1.922,
                        0.171/1.994,
                        1.004/1.191,
                        0.134/1.500,
                        0.420/1.796,
                        1.130/0.131,
                        1.551/0.907,
                        1.049/0.882,
                        0.802/1.119,
                        1.288/0.817,
                        1.432/1.318,
                        0.462/1.344,
                        1.600/0.357,
                        1.305/0.476,
                        1.445/1.711,
                        1.660/0.794,
                        1.336/0.410,
                        0.586/1.793,
                        0.363/1.166,
                        0.843/1.785,
                        1.635/0.684,
                        1.181/0.536,
                        1.248/0.819,
                        1.104/0.872,
                        0.589/1.897,
                        1.527/0.280,
                        1.737/0.975,
                        0.537/1.083,
                        1.267/0.516,
                        0.279/1.670,
                        1.969/1.051,
                        0.037/1.829,
                        0.236/1.153,
                        0.548/1.108,
                        1.303/1.659,
                        0.333/1.476,
                        0.166/1.206,
                        1.438/0.594,
                    } {
                         \node[blue, fill=blue, circle, inner sep=0.45pt, minimum size=1pt] at (\x,\y) {};
                    }
                    \foreach \x/\y in {
                        1.133/0.952,
                        1.327/1.874,
                        1.465/0.430,
                        0.274/1.800,
                        1.748/1.195,
                        1.201/1.330,
                        0.351/1.829,
                        1.038/0.094,
                    } {
                        \node[red, fill=red, regular polygon, regular polygon sides=3, inner sep=0.35pt, minimum size=1pt] at (\x,\y) {};
                    }
                    \node at (-0.7,0.5) {\footnotesize$\mathcal{R}_1$};
                    \draw[->, decorate] 
                     (-0.4,0.5) -- (0.5,0.5);
                    \node at (-0.7,1.5) {\footnotesize$\mathcal{R}_0$};
                    \draw[->, decorate] 
                     (-0.4,1.5) -- (0.5,1.5);
                    \node at (2.6,1.75) {\footnotesize$\mathcal{R}_2$};
                    \draw[->, decorate] 
                     (2.3,1.75) -- (1.7,1.75);
                    \node[
                    align=center,
                    draw=black,      
                    fill=orange!5,             
                    rounded corners,          
                    inner sep=3pt             
                  ] at (4.5,1) {%
                    \footnotesize
                    $\mathcal{H}_0$: Base
                    \footnotesize Heuristic%
                  };
                    \end{tikzpicture}
                  };
                  \node[inner sep=1pt, label={[align=center]below:{(b) Explanation (\eg, Decision Differences: \\ heuristic and optimal route 1–3 differently).
                  }}] (node3) at (5,0) { 
                  \begin{tikzpicture}[>=stealth, node distance=2cm]
                  \label{fig:explanations}
                      \node[draw, circle] (1) at (0,0.75) {1};
                      \node[draw, circle] (2) at (1,0.75) {2};
                      \node[draw, circle] (3) at (2,0.75) {3};
                      \node[draw, circle] (4) at (0,0) {4};
                      \node[draw, circle] (5) at (1,0) {5};
                    
                      \draw[->] (1) -- node[above] {\footnotesize 100} (2);
                      \draw[->] (2) -- node[above] {\footnotesize 100} (3);
                      \draw[->] (1) -- node[left] {\footnotesize 50} (4);
                      \draw[->] (4) -- node[below] {\footnotesize 50} (5);
                      \draw[->] (5) -- (3);
                      \node at ($ (5) + (0.6, 0.25) $) {\footnotesize 50};
                      \draw[->, orange!80, dashed, thick] ($ (1) + (0.2,-0.1) $) -- ($ (3) + (-0.2,-0.1) $);
                      
                      \draw[-, thick, purple!70!black] ($ (1) + (0.1,-0.2) $) -- ($ (4) + (+0.1,+0.15) $);
                      \draw[-, thick, purple!70!black] ($ (4) + (+0.1,+0.15) $) -- ($ (5) + (+0.1,+0.15) $);
                      \draw[->, thick, purple!70!black] ($ (5) + (+0.1,+0.15) $) -- ($ (3) + (-0.2,-0.1) $);

                    \draw[->, thick, set2orange, dashed] (-0.5,-0.6) -- ++(1,0); 
                    \node at (0,-0.4) {{\scriptsize \textcolor{orange!80}{Heuristic}}};

                    \draw[->, thick, purple!70!black] (1,-0.6) -- ++(1,0); 
                    \node at (1.5,-0.4) {{\scriptsize \textcolor{purple!70!black}{Optimal}}};
                      \node[rectangle, draw=none] at (3.3,1) {\footnotesize\textbf{Demand}};
                      \draw (2.5,0.8) -- (4.35,0.8);
                      \node[align=center] at (3,0.65) {\footnotesize{src-dest}};
                      \node at (4,0.65) {\footnotesize{value}};
                      \draw (2.5,0.48) -- (4.35,0.48);
                    
                      \node at (3,0.35) {\footnotesize 1-3};
                      \node at (4,0.35) {\footnotesize 50};
                    
                      \node at (3,0.1) {\footnotesize 1-2};
                      \node at (4,0.1) {\footnotesize 100};
                    
                      \node at (3,-0.15) {\footnotesize 2-3};
                      \node at (4,-0.15) {\footnotesize 100};
                    
                       \draw (2.5,-0.3) -- (4.35,-0.3);
                    
                    \end{tikzpicture}
                  };
                \end{tikzpicture}
            };
            \node[plain] (LLM) at (7.8, 1.5) {\textbf{Suggester LLM}};
            \draw[->, -{Stealth[length=1.75mm]}] (3, 1.5) -- node[above, align=center] {Using $\mathcal{R}_i$ (a) samples \\ and (b) explanations give \\ suggestions to improve $\mathcal{H}_0$} (LLM);

            \node[plain] (generator) at (7.8, -1.5) {\textbf{Heuristic Generator} \\ \textbf{(LLM-Based)}};

            \draw[->, -{Stealth[length=1.75mm]}] (LLM) -- node[left, align=center] {List of suggestions \\ (idea and reasoning)} (generator);
            \draw[->, -{Stealth[length=1.75mm]}, dashed] (generator) -- node[above, align=center] {New heuristic \\ $\mathcal{H}_i$ for region $\mathcal{R}_i$} node[below, align=left] {} (3, -1.5);
        \end{tikzpicture}
    };
    \end{tikzpicture}
      };
    \end{scope}

    \node[align=left, draw=black!50, fill=gray!3, rounded corners,
          font=\footnotesize, text width=4.4cm, drop shadow={shadow xshift=0.5mm, shadow yshift=-0.5mm, opacity=0.5 }]
         at (9, 0) {
      \textbf{Example LLM Suggestions:}\\
      \textbf{1.} "Select a small subset of heavy demands that cross minimal cuts for an exact multi-commodity flow solver over the full graph; fix their routes, then route all other demands on the residual capacities greedily or via the K-best multi-path approach." \\
      \vspace{3pt}
      \textbf{2.} "If a demand’s shortest path includes links whose utilization exceeds a set threshold (\eg, 70\%), trigger a capacity-aware multi-path split: instead of dropping overflow at the bottleneck, inflate saturated link costs to find additional shortest paths and split the flow across them."
    };
  \end{tikzpicture}
  \vspace{-15 pt}
  \caption{\sysname uses heuristic analyzers in a loop with the LLM. The heuristic analyzer finds: (a) regions of the input space where the heuristic underperforms ($\mathcal{R}_1$, $\mathcal{R}_2$); (b) explanations to guide the search. For each region, our solution uses the LLM to suggest how to improve the heuristic for that region and then implements it based on those suggestions. We can run ($\mathcal{H}_1$, $\mathcal{H}_2$) for inputs that come from their corresponding region.\label{fig:architecture}}
 \vspace{-\baselineskip} 
\end{figure*}
It is too much to ask LLMs to design robust heuristics: we show heuristics they generate sometimes underperform (\autoref{fig:funsearch_multiple_subspaces_env}). No matter how cleverly we prompt them, LLMs have a limited circuit size and heuristic design is not a polynomial time problem~\cite{venkat, reactive-synth-lra, reactive-synth-stl}. LLMs, on their own, often cannot infer why and when a heuristic may underperform. This limits their ability to improve the heuristic (\cref{sec:intuition}).

We need heuristics that are resilient to edge-cases, adversarial traffic, and diverse workloads. 
Workloads change over time~\cite{dote}, and the performance of the heuristic on samples from past instances gives little insight about why the heuristic underperformed and often none about how it may perform on new workloads. This makes it hard to produce robust designs. Past research in networking~\cite{soroush, cassini, aifo, packs} is a good indicator that knowing why the heuristics underperform is the key that enables us (humans) to make progress (\cref{sec:explanations}) and it stands to reason the same may hold for LLMs.


We think traditional techniques can help, but we need to enable the LLM to both use them and to interpret their outputs. This is hard.

It would help to provide the LLM with information about why the heuristic underperformed in each case. We need to convert these ``explanations'' into a form that benefits the LLM-based search: we find it is better to first use the LLM to distill these insights into ``suggestions'' on how to improve the heuristic (see~\autoref{fig:oneshot}).



Many heuristics are approximate solutions to NP-hard problems. So, no matter how cleverly we prompt the LLM, a general polynomial-time heuristic that would perform well in the corner cases may not exist. We hypothesize it may be easier to produce an ensemble of heuristics, each specialized to regions of the input space, instead of a generalist that would perform well across the entire space. We tried this idea: we found ``good'' subregions of the input space and used the LLM to design specialized strategies for each one~---~the heuristics the LLM produced were more robust (see~\cref{sec:ensemble}) compared to the unmodified process. 
It is hard to produce good regions in general because it depends on the problem structure and impacts the quality of the result.



Some of these problems are easier to tackle in the networking domain: (1) we have well-defined protocols that standardize the network's behavior~---~we can leverage the predictability this brings to ``simplify'' the problem; (2) we have a network topology in which we see ``repeated'' behaviors which we may be able to use to scale our solutions. This intuition is the same that has enabled us to scalably apply techniques from programming languages to networking problems in network verification (\eg,~\cite{minaQueue, ccac, batfish, yu}). We expand on these opportunities as we discuss open questions.

\noindent \textbf{Summary.} We introduce \sysname (\autoref{fig:architecture}), a new solution to LLM-based heuristic design; novel mechanisms that allow us to address some of the challenges we discussed above; and propose research directions that help address the others. The design starts with a base-heuristic and uses heuristic analyzers alongside calls to an LLM to partition the space into regions; ``explain'' why the heuristic underperforms; and derive ``suggestions'' on how to improve it. It then modifies the search to (1) create a specialized heuristic for each region; and (2) use the suggestions to find better heuristics.
Our initial results show the mechanisms we introduce are viable.  

Our contributions are as follows:
\begin{compactitem}
\item We propose a novel architecture for LLM-based heuristic design through explanation-guided genetic search.
\item We use a case study of a traffic engineering problem to show these mechanisms are viable. We introduce certain design principles and discuss techniques that we found ineffective. We show \sysname finds heuristics that are $28\times$ better than the baselines.
\item We discuss open questions and the opportunity to solve them. We also describe how we may solve these challenges and why they may be more tractable problems to solve in the networking space. 
\end{compactitem}

For our initial prototype, we use existing heuristic analyzers and algorithms that explain heuristic performance, such as MetaOpt~\cite{metaopt} and XPlain~\cite{xplain}, but our architecture is general and can support any similar tool.

\section{Background}
\label{sec:background}

Researchers and practitioners have used LLMs for program synthesis. Many copilots provide an interactive framework to help write programs~\cite{copilot-eval,copilot-main, codex, jules, curser}, sometimes with human guidance~\cite{llm-prog-synth}. Others generate tests to enable LLMs to critique their own outputs~\cite{alpha-code, testgeneration, selfReflection}. Researchers have used formal methods to guide LLM-based synthesis across domains such as program generation~\cite{synver,dreamcoder}, loop-invariant inference~\cite{llm-bmc}, and network configuration management~\cite{GeorgeLLM}.

Recent work on LLM-based heuristic design presents a new breakthrough~\cite{funsearch, gao2024alpha, liu2024evolution, he2024designing, yang2025heuragenix, AlphaVerus}. We discuss FunSearch~\cite{funsearch} as an example. FunSearch uses genetic search algorithms where the LLM does cross-over (combines different heuristics to create stronger ones) and mutation (introduces random changes to the heuristic to explore the space).

FunSearch's algorithm starts with a~\emph{set} of base heuristics. These heuristics are the initial seed to ``clusters'' which the algorithm evolves over time.
Users provide an ``evaluation function'' that scores the heuristics based on the heuristic's performance on random samples from the input space.

FunSearch iterates as follows: (1) it evolves the heuristics within each cluster for a pre-specified amount of time; (2) it then evaluates the heuristics in each cluster and assigns a score to the cluster based on the best performing heuristic in it; (3) the algorithm then removes half of the lowest scoring clusters and spawns new ones to replace them. To seed the new clusters, the algorithm uses the best heuristic from the clusters that survived the previous round. The algorithm terminates after a pre-set timeout.


\section{The promise and open questions}
\label{sec:intuition}

LLM-based program synthesis differs from FunSearch---and from our focus---in an important way: synthesized programs are typically judged as either correct or incorrect. Heuristics, by contrast, lack such a binary criterion. Their evaluation is based on performance, which is not only quantitative but often nuanced. Which parts of the input space matter? Should we judge the heuristic based on its average performance? on the tail performance? or the worst case? What input distributions should we use to evaluate the heuristic? should we focus on specific input instances or on the entire space?

If we measure its average performance, then we may potentially mask the heuristic's poor performance on rare (but critical) inputs because of its strong performance elsewhere. This is especially true when we evaluate the heuristic on arbitrary or random samples from the input space. 


A single counter-example rarely captures the full extent of a heuristic's weaknesses. A heuristic's performance often suffers under many different ``types'' of inputs, which makes it harder to diagnose and improve. Many of these heuristics try to find polynomial-time approximations for NP-hard problems or faster heuristics for polynomial-time problems (\eg, in traffic engineering). Often, no single fast heuristic can perform well across the entire input space~---~we are more likely to succeed with specialized heuristics that combine multiple complementary strategies.



We next discuss these concepts in depth, show how they help generate better heuristics through a concrete example, and discuss open problems our community needs to solve in order to apply these techniques in practical settings.

\subsection{Adversarial samples or random ones?}
\label{subsec:adversarial}

FunSearch and similar algorithms evaluate heuristics they generate on random samples over the input space and use the result as feedback to guide the search. But such random samples~---~especially those that come from a uniform (or other ad-hoc) distributions~---~can miss important regions of the input space. A heuristic may perform well on most of these inputs, yet underperform on important practical instances. Such instances occasionally do happen in real workloads and cause severe consequences~\cite{arun2022starvation,raha,metaopt}.


\noindent\textbf{Promise.} If we evaluate the average case performance of a heuristic, we can mask the scenarios in which the heuristic underperforms, especially if these occur in small pockets of the input space. It is not easy for a user to modify the sampling distribution to fix this issue (to increase the probability that the evaluator draws samples from regions where the heuristic may underperform) because they may not be able to identify when, where, and how they should do so. But we hypothesize if we bridge this gap we can then guide the synthesis process towards more robust solutions.



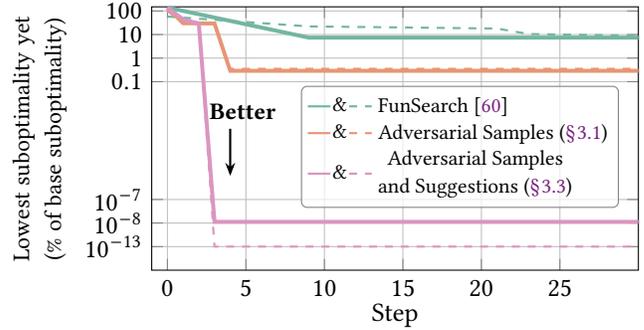
\begin{figure}[t]
    \centering
    \begin{tikzpicture}
    \begin{semilogyaxis}[
        width=0.95\columnwidth,
        height=0.6\columnwidth,
        xlabel={Step},
        xlabel style={yshift=4pt},
        ylabel={\small Lowest suboptimality yet\\ \small(\% of base suboptimality)},
        ylabel style={align=center},
        legend style={anchor=north east, legend                 columns=1,font=\footnotesize,
            fill=white,
            at={(0.96,0.7)}, 
            draw opacity=1,
            text opacity=1,
            fill opacity=0.7,
            rounded corners=.075cm,
            inner sep=0.07cm,
            row sep=-1pt,
            draw=gray},
        grid=both,
        grid=major,
        ymin=10e-10,
        ymax=150,
        xmin=-1,
        xmax=30,
        xtick={0,5,10,15,20,25},
        y tick style={      
            line width=0.2pt,
        },
        ytick={10e-9, 10e-8, 10e-7, 0.1, 1, 10, 100},
        yticklabels={$10^{-13}$, $10^{-8}$, $10^{-7}$, $0.1$, $1$, $10$, $100$},
        legend cell align=left,
    ]
    
    \addplot[dashed, color=set2cyan, thick, forget plot] coordinates {
(0.0, 57.983392864) (9.0, 22.165934678) (21.0, 17.978958011) (23.0, 10.549309772) (35.0, 7.826953658) (50.0, 7.826953658)
    };

    
    \addplot[solid, color=set2cyan, line width=1.5pt, forget plot] coordinates {
 (0, 139.98738153761454) (9, 7.415018048407125) (10, 7.415018048407125) (11, 7.415018048407125) (12, 7.415018048407125) (13, 7.415018048407125) (14, 7.415018048407125) (15, 7.415018048407125) (16, 7.415018048407125) (17, 7.415018048407125) (18, 7.415018048407125) (19, 7.415018048407125) (20, 7.415018048407125) (21, 7.415018048407125) (22, 7.415018048407125) (23, 7.415018048407125) (24, 7.415018048407125) (25, 7.415018048407125) (26, 7.415018048407125) (27, 7.415018048407125) (28, 7.415018048407125) (29, 7.415018048407125) (30, 7.415018048407125)
    };

    
    \addplot[dashed, color=set2orange, thick, forget plot] coordinates {
(0, 99.99999977469972) (1, 30.041605717661017) (3, 29.28738495338234) (4, 0.3510085618781723) (5, 0.3510085618781723) (6, 0.3510085618781723) (7, 0.3510085618781723) (8, 0.3510085618781723) (9, 0.3510085618781723) (10, 0.3510085618781723) (11, 0.3510085618781723) (12, 0.3510085618781723) (13, 0.3510085618781723) (14, 0.3510085618781723) (15, 0.3510085618781723) (16, 0.3510085618781723) (17, 0.3510085618781723) (18, 0.3510085618781723) (19, 0.3510085618781723) (20, 0.3510085618781723) (21, 0.3510085618781723) (22, 0.3510085618781723) (23, 0.3510085618781723) (24, 0.3510085618781723) (25, 0.3510085618781723) (26, 0.3510085618781723) (27, 0.3510085618781723) (28, 0.3510085618781723) (29, 0.3510085618781723) (30, 0.3510085618781723)
    };
    
    \addplot[solid, color=set2orange, line width=1.5pt, forget plot] coordinates {
(0, 139.98738153761454) (1, 30.14242030473075) (3, 29.438054783479075) (4, 0.2888376017041432) (5, 0.2888376017041432) (6, 0.2888376017041432) (7, 0.2888376017041432) (8, 0.2888376017041432) (9, 0.2888376017041432) (10, 0.2888376017041432) (11, 0.2888376017041432) (12, 0.2888376017041432) (13, 0.2888376017041432) (14, 0.2888376017041432) (15, 0.2888376017041432) (16, 0.2888376017041432) (17, 0.2888376017041432) (18, 0.2888376017041432) (19, 0.2888376017041432) (20, 0.2888376017041432) (21, 0.2888376017041432) (22, 0.2888376017041432) (23, 0.2888376017041432) (24, 0.2888376017041432) (25, 0.2888376017041432) (26, 0.2888376017041432) (27, 0.2888376017041432) (28, 0.2888376017041432) (29, 0.2888376017041432) (30, 0.2888376017041432)
    };
    
    \addplot[dashed, color=set2purple, thick,forget plot] coordinates {
(0, 99.99999977469972) (1, 44.48242356832569) (2, 29.47329199718085) (3, 10e-9) (4, 10e-9) (5, 10e-9) (6, 10e-9) (7, 10e-9) (8, 10e-9) (9, 10e-9) (10, 10e-9) (11, 10e-9) (12, 10e-9) (13, 10e-9) (14, 10e-9) (15, 10e-9) (16, 10e-9) (17, 10e-9) (18, 10e-9) (19, 10e-9) (20, 10e-9) (21, 10e-9) (22, 10e-9) (23, 10e-9) (24, 10e-9) (25, 10e-9) (26, 10e-9) (27, 10e-9) (28, 10e-9) (29, 10e-9) (30, 10e-9)
    };

    \addplot[solid, color=set2purple, line width=1.5pt, forget plot] coordinates {
 (0, 139.98738153761454) (1, 43.66131247992361) (2, 29.399844266860057) (3, 1.1234906456734554e-07) (4, 1.1234906456734554e-07) (5, 1.1234906456734554e-07) (6, 1.1234906456734554e-07) (7, 1.1234906456734554e-07) (8, 1.1234906456734554e-07) (9, 1.1234906456734554e-07) (10, 1.1234906456734554e-07) (11, 1.1234906456734554e-07) (12, 1.1234906456734554e-07) (13, 1.1234906456734554e-07) (14, 1.1234906456734554e-07) (15, 1.1234906456734554e-07) (16, 1.1234906456734554e-07) (17, 1.1234906456734554e-07) (18, 1.1234906456734554e-07) (19, 1.1234906456734554e-07) (20, 1.1234906456734554e-07) (21, 1.1234906456734554e-07) (22, 1.1234906456734554e-07) (23, 1.1234906456734554e-07) (24, 1.1234906456734554e-07) (25, 1.1234906456734554e-07) (26, 1.1234906456734554e-07) (27, 1.1234906456734554e-07) (28, 1.1234906456734554e-07) (29, 1.1234906456734554e-07) (30, 1.1234906456734554e-07)
    };

    \draw[-{Stealth[length=1.75mm]}, thick] (axis cs:4,0.001) -- (axis cs:4,0.00001);
    \node[anchor=south east, font=\small] at (axis cs:7.5,0.001) {\textbf{Better}};

    \addlegendimage{legend image code/.code={
         \draw[set2cyan, thick] (0cm,0cm) -- (0.3cm,0cm);
        \node at (0.25cm, 0cm) {\&};
        \draw[set2cyan, thick, dashed] (0.55cm,0cm) -- (0.85cm,0cm);
    }}
    \addlegendentry{FunSearch~\cite{funsearch}}

    \addlegendimage{legend image code/.code={
        \draw[set2orange, thick] (0cm,0cm) -- (0.3cm,0cm);
        \node at (0.25cm, 0cm) {\&};
        \draw[set2orange, thick, dashed] (0.55cm,0cm) -- (0.85cm,0cm);
    }}
    \addlegendentry{Adversarial Samples~(\cref{subsec:adversarial})}

    \addlegendimage{legend image code/.code={
        \draw[set2purple, thick] (0cm,0cm) -- (0.3cm,0cm);
        \node at (0.25cm, 0cm) {\&};
        \draw[set2purple, thick, dashed] (0.55cm,0cm) -- (0.85cm,0cm);
    }}
    \addlegendentry{\shortstack{Adversarial Samples \\ and Suggestions (\cref{sec:explanations})}}
    \end{semilogyaxis}

    \end{tikzpicture}
    \vspace{-0.8cm}
    \caption{We compare FunSearch with solutions where we (1) focus on adversarial samples in each step (\cref{subsec:adversarial}); and (2) also incorporate  ``suggestions'' in each step (\cref{sec:explanations}); on a traffic engineering problem. Dashed lines report the lowest (best) suboptimality achieved for a heuristic on training samples so far, and the solid lines measure it for a held-out set of adversarial samples. The suboptimality is the worst-case performance of a heuristic compared to the optimal across all samples.} 
  \label{fig:funsearch_env}
   \vspace{-\baselineskip} 
\end{figure}

Our preliminary results support this hypothesis. In experiments where we targeted Microsoft’s traffic engineering heuristic~\cite{oneWAN,demandpinning}---which ``pins'' small demands to their shortest paths and optimally routes the rest---we expose the LLM to \emph{adversarial inputs} (input instances that cause the largest performance problems for the heuristic) and their \emph{individual performance suboptimalities} (the performance gap between the heuristic and the optimal solution)\footnote{We define the (overall) \emph{suboptimality} as the worst-case suboptimalities across all input instances}. We found that even this simple signal allowed FunSearch to create heuristics that outperformed (they had a lower suboptimality) what it found before (\autoref{fig:funsearch_env}). We describe the experiment details in \cref{sec:eval}.

\noindent \textbf{Open questions.} 
Ideally, we would revise the adversarial inputs at each iteration of the search process, since they evolve alongside the heuristic. But most tools that produce adversarial inputs require a mathematical model of the heuristic~\cite{metaopt, minaQueue,venkat}. We found LLMs, out of the box, could not provide such models. Thus, in our experiments, we generated the adversarial instances only for the base heuristic and reused them to evaluate all subsequent heuristics during the search. Even this limited signal was valuable.


Prior work has used LLMs to produce mathematical models for certain problems~\cite{optiguide,li2024towards}~---~we have reason to be optimistic. Several domain-specific languages (DSLs) already exist to make such modeling accessible---for example, MiniZinc~\cite{minizinc} and AMPL~\cite{ampl}. XPlain~\cite{xplain} introduces a DSL that helps model heuristics as optimization problems. These DSLs may enable LLMs to model heuristics automatically. 

Another challenge is scalability. SMT-based~\cite{ccac, venkat, minaQueue} and optimization-based heuristic analyzers~\cite{metaopt,xplain} are slow. This is acceptable for offline analysis of a single heuristic, but it becomes a bottleneck in search-driven synthesis frameworks that need to quickly iterate over many heuristics. We can mitigate this problem with domain-specific strategies. For example, we can use partitioning techniques inspired by NCFlow~\cite{ncflow} and POP~\cite{pop} to scale heuristic analysis for wide-area networks~\cite{metaopt}; or graph isomorphism for those that target data centers~\cite{propane, janus,oblivious}; or use specialized encodings to analyze queues or schedulers~\cite{minaQueue, metaopt}. But we need more research to automatically find and apply the appropriate scaling strategy (and maybe even more than one).



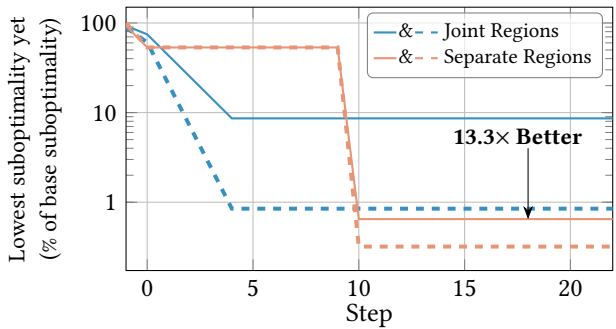
\begin{figure}[t]
    \centering
    \begin{tikzpicture}
    \begin{semilogyaxis}[
        width=0.95\columnwidth,
        height=0.6\columnwidth,
        xlabel={Step},
        xlabel style={yshift=4pt},
        ylabel={\small Lowest suboptimality yet \\ \small(\% of base suboptimality)},
        ylabel style={align=center},
        legend style={anchor=north east, legend                 columns=1,font=\footnotesize,
            fill=white,
            draw opacity=1,
            text opacity=1,
            fill opacity=0.7,
            rounded corners=.075cm,
            inner sep=0.07cm,
            row sep=-1pt,
            draw=gray},
        grid=both,
        grid=major,
        ymin=0,
        ymax=150,
        xmin=-1,
        xmax=22,
        xtick={0,5,10,15,20},
        y tick style={      
            line width=0.2pt,
        },
        ytick={1, 10, 100},
        yticklabels={$1$, $10$, $100$},
        legend cell align=left,
    ]
    
    \addplot[dashed, tealblue, line width=1.5pt, forget plot] coordinates {
    (-1,88.598) (0,60.777) (4,0.8474) (5,0.8474) (6,0.8474) (7,0.8474) (8,0.8474)
    (9,0.8474) (10,0.8474) (11,0.8474) (12,0.8474) (13,0.8474) (14,0.8474)
    (15,0.8474) (16,0.8474) (17,0.8474) (18,0.8474) (19,0.8474) (20,0.8474)
    (21,0.8474) (22,0.8474)
    };
    
    \addplot[solid, tealblue, thick, forget plot] coordinates {
    (-1,92.931) (0,74.682) (4,8.6209) (5,8.6209) (6,8.6209) (7,8.6209) (8,8.6209)
    (9,8.6209) (10,8.6209) (11,8.6209) (12,8.6209) (13,8.6209) (14,8.6209)
    (15,8.6209) (16,8.6209) (17,8.6209) (18,8.6209) (19,8.6209) (20,8.6209)
    (21,8.6209) (22,8.6209)
    };
    
    \addplot[dashed, set2orange, line width=1.5pt, forget plot] coordinates {
    (-1,100.0) (0,53.346) (2,53.346) (5,53.346) (7,53.346) (9,53.346)
    (10,0.3198) (16,0.3198) (22,0.3198)
    };
    
    \addplot[solid, set2orange, thick, forget plot] coordinates {
    (-1,92.931) (0,53.333) (2,53.333) (5,53.333) (7,53.333) (9,53.333)
    (10,0.6482) (16,0.6482) (22,0.6482)
    };

    \addlegendimage{legend image code/.code={
        \draw[tealblue, thick] (0cm,0cm) -- (0.3cm,0cm);
        \node at (0.25cm, 0cm) {\&};
        \draw[tealblue, thick, dashed] (0.55cm,0cm) -- (0.85cm,0cm);
    }}
    \addlegendentry{Joint Regions}

    \addlegendimage{legend image code/.code={
        \draw[set2orange, thick] (0cm,0cm) -- (0.3cm,0cm);
        \node at (0.25cm, 0cm) {\&};
        \draw[set2orange, thick, dashed] (0.55cm,0cm) -- (0.85cm,0cm);
    }}
    \addlegendentry{Separate Regions}

    \draw[-{Stealth[length=1.75mm]}] (axis cs:18,4) -- (axis cs:18,0.648);
    \node[anchor=south east, font=\small] at (axis cs:21,3.5) {\textbf{13.3$\times$ Better}};
    
    \end{semilogyaxis}
    \end{tikzpicture}
    \vspace{-0.5cm}
    \caption{we create an approach which uses FunSearch to create specialized heuristics for each region (separate regions) with one that generates a single heuristic to operate over the entire space. We find specialized heuristics outperform monolithic ones. The dashed and solid lines show overall suboptimality so far on training and held-out samples, respectively.}
  \label{fig:funsearch_multiple_subspaces_env}
   \vspace{-\baselineskip} 
\end{figure}

\subsection{Ensemble of heuristics or just one?}
\label{sec:ensemble}
Inputs in different regions of the input space often exhibit shared structural properties which a specialized heuristic can exploit. An ensemble of specialists is likely to outperform generalists that FunSearch-like techniques produce.


\noindent \textbf{Promise.} We use XPlain~\cite{xplain} to partition the input space into regions where the base heuristic underperforms. We then compare the performance of two approaches: one where FunSearch generates a single general-purpose heuristic, and another where it synthesizes a separate heuristic for each region, which we then combine into an ensemble (\autoref{fig:funsearch_multiple_subspaces_env}).

In our experiments, one region of the input space consisted of scenarios with a few large demands that shared bottlenecks with many smaller flows. The base heuristic, which ``pins'' small demands to their shortest paths, performed poorly here: it pre-committed small flows to a route and left insufficient capacity for larger demands. The specialized heuristic reversed this order: it sorted demands by size and allocated capacity to larger flows before smaller ones.

While the heuristic above may not always perform well (it is slower since it does not remove as many demands from the optimization as the base and since it prioritizes large demands its mistakes are more costly) but it improves performance~\emph{in this region}.
Microsoft deployed the ``pinning'' heuristic because such skewed demands are less common. But recent work has shown when such demands happen in practice they can degrade performance by $30\%$~\cite{metaopt}. Through this specialized heuristic, we can have the best of both worlds and switch to the specialized heuristic when we see demands that fit this pattern.

Operators can either inspect the demand in each iteration to select an appropriate heuristic, or run multiple heuristics in parallel and pick the best.

\noindent \textbf{Open questions.} We used regions that trigger a similar behavior in the base heuristic~\cite{xplain}. But there are other viable approaches. For example, symbolic execution tools such as Klee~\cite{klee} allow us to find equivalence classes of inputs that trigger the same code-paths in an algorithm~---~we can view each such code path in the optimal algorithm, the base heuristic, or the Cartesian product of both as a new region.

There is an opportunity to define these regions in a domain-specific way and based on practical insights. For example, graph-partitioning algorithms~\cite{nazi2019gap, brucker1978complexity} generate good equivalence classes for traffic engineering heuristics. We need to develop techniques to select the best approach automatically.



Our approach is expensive at larger scales. We hypothesize that the principle of bounded model checking may help. It may be possible to (1) derive the regions for lower-dimensional inputs and then project them into regions in the higher-dimensional space; and/or (2) find ``concepts'' that describe how to improve the heuristics based on inputs in lower-dimensions and then apply those concepts in the higher-dimensional space (this is somewhat similar to~\cite{sangeethaConcept}); and/or (3) express partitions in a ``scale-invariant'' DSL. The LLM itself may help enable such solutions. 

We have some evidence that this may work. We used our approach and a medium sized (see \cref{sec:eval}) topology to generate more robust heuristics and then evaluated the heuristics we generated on a large one (one with $196$ nodes and $486$ edges). The worst suboptimality (as determined by running a heuristic analyzer on the larger topology) was $ 46\%$ on the larger topology when we ran the base heuristic. The heuristic we created based on the smaller topology, without further modification, reduced this to $27\%$.

We defined regions statically (we define them once and we analyze the base heuristic to do so). The base heuristic we start with may not be a good one~---~this heuristic just seeds the process and is only meant as a starting point for the search. It may be unwise to define the regions based on this heuristic. 
We can solve this problem if we instead define the regions based on the optimal algorithm instead (but this can be too slow in most cases) or if we re-evaluate the regions as we find better heuristics during the search. How and when to apply each approach is an open problem.

{\small
\begin{figure}
    \centering
    \begin{tikzpicture}
    \begin{axis}[
        width=1\columnwidth,
        height=0.6\columnwidth,
        ymode=log,
        boxplot/draw direction=y,
        ylabel={\shortstack{\small Suboptimality \\ \small(\% of base suboptimality)}},
        xlabel style={yshift=2pt},
        ylabel style={yshift=-3pt},
        xtick={1,2,3,4},
        ymin=20,
        ytick={20, 50, 100, 200, 500},
        log ticks with fixed point,
        xticklabels={
            {Vanilla},
            {Samples},
            {Samples\\ \& Explanations},
            {Suggestions}
        },
        grid=both,
        ymajorgrids=true,
        xmajorgrids=false,
        xtick align=outside,
        xtick pos=bottom,
        x tick style={      
            color=black,      
            line width=0.6pt,
        },
        tick label style={font=\small, align=center},
        ytick style={draw=none},
        major tick length=1mm,
        grid style={dashed, lightgray, dash pattern=on 1pt off 1pt},
    ]
    
        \addplot+[
            boxplot prepared={
                median=109.64/210*100,
                average=323.03/210*100,
                upper quartile=597.58/210*100,
                lower quartile=99.61/210*100,
                upper whisker=657.58/210*100,
                lower whisker=99.61/210*100
            },
            fill=set2cyan,
            draw=set2gray,
            line width=0.8pt
        ] coordinates {};
    
        \addplot+[
            boxplot prepared={
                average=282.65/210*100,
                median=104.65/210*100,
                upper quartile=113.96/210*100,
                lower quartile=99.62/210*100,
                upper whisker=115.39/210*100,
                lower whisker=93.26/210*100
            },
            fill=set2orange,
            draw=set2gray,
            line width=0.8pt
        ] coordinates {};
    
        \addplot+[
            boxplot prepared={
                average=102.99/210*100,
                median=105.75/210*100,
                upper quartile=106.67/210*100,
                lower quartile=99.61/210*100,
                upper whisker=109.64/210*100,
                lower whisker=93.26/210*100
            },
            fill=set2blue,
            draw=set2gray,
            line width=0.8pt
        ] coordinates {};
    
        \addplot+[
            boxplot prepared={
                average=79.0/210*100,
                median=81.2/210*100,
                upper quartile=90.65/210*100,
                lower quartile=68.85/210*100,
                upper whisker=99.64/210*100,
                lower whisker=53.19/210*100
            },
            fill=set2purple,
            draw=set2gray,
            line width=0.8pt
        ] coordinates {};

        \addplot+[
                only marks, draw=set2gray, fill=none, mark=o, mark size=2.5, line width=0.8pt
            ] coordinates {
            (2, {1178.35/210*100})
            };
        \draw[-{Stealth[length=1.75mm]}, thick] (axis cs:4,250) -- (axis cs:4,120);
        \node[align=center, anchor=west] at (axis cs:3.6,300) {\small \textbf{Better}};
    \end{axis}
    \end{tikzpicture}
    \vspace{-1.75em}
    \caption{Explanations benefit the search. We consider one step of the search process and see that we can find better heuristics if we first ``explain'' why a heuristic underperforms. We can improve the approach further if we convert these explanations into suggestions on how to improve the heuristic. We evaluate all of the approaches on a held-out set of adversarial samples.}
    \label{fig:oneshot}
     \vspace{-\baselineskip}
\end{figure}
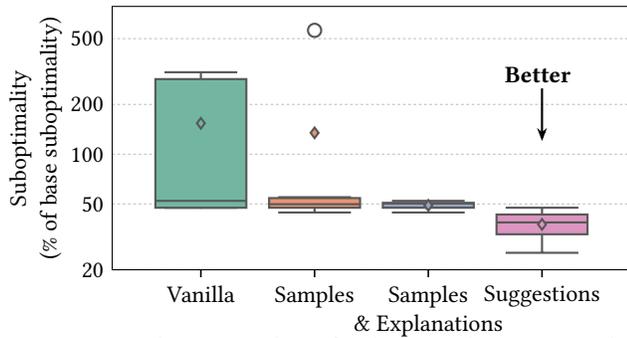
}

\subsection{Do explanations help?}
\label{sec:explanations}

Researchers have worked to improve networking heuristics for decades. SWAN~\cite{swan} designed an algorithm that out-performed the traffic engineering solution Danna \ETAL~\cite{danna} had proposed and Soroush~\cite{soroush} then design an algorithm that improved it further. Cassini~\cite{cassini} improves Themis~\cite{themis}. TACCL~\cite{taccl} improves MSCCL~\cite{msccl} which is itself outperformed by~\cite{liangyu}. PACKs~\cite{packs} designs a new packet scheduling algorithm that improves upon SP-PIFO~\cite{sp-pifo} and AIFO~\cite{aifo}.

Most of these works identify~\emph{why} a heuristic underperforms and address that root cause directly. For example, to improve max-min fair resource allocation algorithms, researchers found the core difference between the heuristics and the optimal was in the heuristics' ability to approximately sort demands based on how much capacity the algorithm assigned to them~\cite{soroush,danna,swan}. Each new solution targeted this root cause and achieved better and better approximations more quickly.  We find a similar methodology in the novel heuristics designed by the MetaOpt authors~\cite{metaopt} for the demand-pinning heuristic in traffic engineering and for the SP-PIFO packet scheduling algorithms. The CCAC authors~\cite{venkat} also did the same for congestion control. We thus think explanations are the key to better heuristic design.

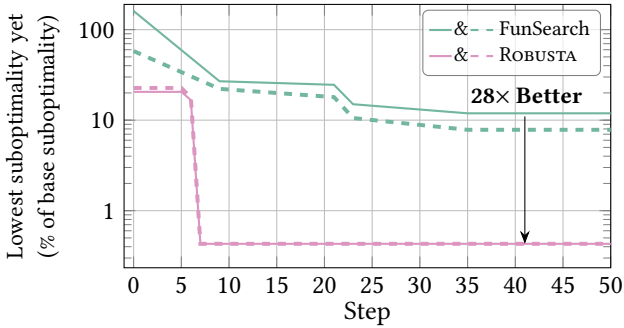
\begin{figure}[t]
    \centering
    \begin{tikzpicture}
    \begin{semilogyaxis}[
        width=0.95\columnwidth,
        height=0.6\columnwidth,
        xlabel={Step},
        xlabel style={yshift=4pt},
        ylabel={\small Lowest suboptimality yet\\ \small (\% of base suboptimality)},
        ylabel style={align=center},
        legend style={anchor=north east, legend                 columns=1,font=\footnotesize,
            fill=white,
            draw opacity=1,
            text opacity=1,
            fill opacity=0.7,
            rounded corners=.075cm,
            inner sep=0.07cm,
            row sep=-1pt,
            draw=gray},
        grid=both,
        grid=major,
        ymin=0,
        ymax=190,
        xmin=-1,
        xmax=50,
        xtick={0,5,10,15,20, 25, 30, 35, 40, 45, 50},
        y tick style={      
            line width=0.2pt,
        },
        ytick={0.1, 1, 10, 100},
        yticklabels={$0.1$, $1$, $10$, $100$},
        legend cell align=left,
    ]
    
    \addplot[dashed, set2cyan, line width=1.5pt, forget plot] coordinates {
(0.0, 57.983392864) (9.0, 22.165934678) (21.0, 17.978958011) (23.0, 10.549309772) (35.0, 7.826953658) (50.0, 7.826953658)
    };
    
    \addplot[solid, set2cyan, thick, forget plot] coordinates {
 (0.0, 160.786441351) (9.0, 26.912319922) (21.0, 24.576541923) (23.0, 14.999059785) (35.0, 11.903413890) (50.0, 11.903413890)
    };
    
    \addplot[dashed, set2purple, line width=1.5pt, forget plot] coordinates {
(0, 22.681589377) (1, 22.681589377) (2, 22.681589377) (3, 22.681589377) (4, 22.681589377) (5, 22.681589377) (6, 17.806179405) (7, 0.430422490) (8, 0.430422490) (9, 0.430422490) (10, 0.430422490) (11, 0.430422490) (12, 0.430422490) (13, 0.430422490) (14, 0.430422490) (15, 0.430422490) (16, 0.430422490) (17, 0.430422490) (18, 0.430422490) (19, 0.430422490) (20, 0.430422490) (21, 0.430422490) (22, 0.430422490) (23, 0.430422490) (24, 0.430422490) (25, 0.430422490) (26, 0.430422490) (27, 0.430422490) (28, 0.430422490) (29, 0.430422490) (30, 0.430422490) (31, 0.430422490) (32, 0.430422490) (33, 0.430422490) (34, 0.430422490) (35, 0.430422490) (36, 0.430422490) (37, 0.430422490) (38, 0.430422490) (39, 0.430422490) (40, 0.430422490) (41, 0.430422490) (42, 0.430422490) (43, 0.430422490) (44, 0.430422490) (45, 0.430422490) (46, 0.430422490) (47, 0.430422490) (48, 0.430422490) (49, 0.430422490) (50, 0.430422490)
    };
    
    \addplot[solid, set2purple, thick, forget plot] coordinates {
 (0, 20.542371814) (1, 20.542371814) (2, 20.542371814) (3, 20.542371814) (4, 20.542371814) (5, 20.542371814) (6, 16.685410694) (7, 0.430422490) (8, 0.430422490) (9, 0.430422490) (10, 0.430422490) (11, 0.430422490) (12, 0.430422490) (13, 0.430422490) (14, 0.430422490) (15, 0.430422490) (16, 0.430422490) (17, 0.430422490) (18, 0.430422490) (19, 0.430422490) (20, 0.430422490) (21, 0.430422490) (22, 0.430422490) (23, 0.430422490) (24, 0.430422490) (25, 0.430422490) (26, 0.430422490) (27, 0.430422490) (28, 0.430422490) (29, 0.430422490) (30, 0.430422490) (31, 0.430422490) (32, 0.430422490) (33, 0.430422490) (34, 0.430422490) (35, 0.430422490) (36, 0.430422490) (37, 0.430422490) (38, 0.430422490) (39, 0.430422490) (40, 0.430422490) (41, 0.430422490) (42, 0.430422490) (43, 0.430422490) (44, 0.430422490) (45, 0.430422490) (46, 0.430422490) (47, 0.430422490) (48, 0.430422490) (49, 0.430422490) (50, 0.430422490)
    };

    \addlegendimage{legend image code/.code={
        \draw[set2cyan, thick] (0cm,0cm) -- (0.3cm,0cm);
        \node at (0.25cm, 0cm) {\&};
        \draw[set2cyan, thick, dashed] (0.55cm,0cm) -- (0.85cm,0cm);
    }}
    \addlegendentry{FunSearch}

    \addlegendimage{legend image code/.code={
        \draw[set2purple, thick] (0cm,0cm) -- (0.3cm,0cm);
        \node at (0.25cm, 0cm) {\&};
        \draw[set2purple, thick, dashed] (0.55cm,0cm) -- (0.85cm,0cm);
    }}
    \addlegendentry{\sysname}

    \draw[-{Stealth[length=1.75mm]}] (axis cs:41,11) -- (axis cs:41, 0.430422490);
    \node[anchor=south east, font=\small] at (axis cs:48,12) {\textbf{28$\times$ Better}};
    
    \end{semilogyaxis}
    \end{tikzpicture}
    \vspace{-0.5cm}
    \caption{We evaluate our ideas end-to-end. We create specialized heuristics on $5$ regions and use approximate suggestions. \sysname finds heuristics with better worst-case performance.}
  \label{fig:funsearch_e2e}
  \vspace{-10pt}
\end{figure}

\noindent\textbf{Promise.}~\cref{fig:oneshot} confirms this. We evaluate whether explanations improve the quality of the heuristic the LLM produces in each step of the search (we can think of this as a single-shot experiment where we only run one step of the search). We show the result of the full (where we approximate the explanations in each step) search in~\autoref{fig:funsearch_e2e}.

The vanilla approach prompts the LLM to improve the base heuristic. The ``samples'' approach changes the prompt to include samples that cause the heuristic to underperform (this captures the idea that ``adversarial samples'' can help design better heuristics). The ``samples and explanations'' approach feeds the LLM decision differences (similar to that in prior work~\cite{xplain, yang2025heuragenix}) that characterizes where the heuristic and the optimal took different actions for the same inputs in the sample-set. The suggestions approach first converts the samples and explanations into ``suggestions'' (through calls to the LLM) for how to improve the heuristic; feeds the suggestions into the LLM one at a time to create new heuristics; and returns the best one. Each approach prompts the LLM ten times to create a new heuristic and reports the best performance across these ten heuristics.



Suggestions are the most effective. This is because, if we directly include the samples and explanations, we increase the length of the prompt  which degrades the LLM's performance (it is well-known that LLM's performance degrades on longer contexts~\cite{longcontext1, longcontext2, longcontext3}). Suggestions summarize the relevant information in these inputs and shorten the context.

Higher dimensional samples obscure the root cause for LLMs and humans alike and degrade performance further. Our results show the baseline improves on lower-dimensional samples. But we need large enough instances so that we can trigger the mechanisms that cause the heuristic to underperform (some problems may only manifest at large~\emph{enough} dimensions). Suggestions may help summarize such instances. 



\noindent \textbf{Open questions.} The above points to an interesting property. \emph{If} the principles of bounded model checking apply \emph{then} it is better to first find the smallest instance where we can observe the heuristic's performance problems and try to improve it with samples and explanations we find in that scale. This is hard as the right scale can depend on the heuristic itself and other aspects of the problem.

How can we explain why a heuristic underperforms? Prior works show initial steps towards such explanations~\cite{xplain, minaQueue, ccac, arun2022starvation}, but none can analyze the code the LLM generates.

While we show suggestions help FunSearch, it remains an open question (once we find a way to automatically generate suggestions to improve new heuristics) how to use these suggestions throughout the search and whether updated suggestions can result in further gains (in our evaluations in \cref{sec:eval} we generate approximate explanations but exact explanations can help improve the results even more). The LLM often suggests multiple mechanisms for how we can improve the heuristic, we currently use each one to start the search but it remains open how to use them \emph{during} the search to either create new clusters (\cref{sec:background}) or update existing ones.

\section{Evaluation}
\label{sec:eval}
We propose \sysname, that incorporates these ideas (\autoref{fig:architecture}). In each step, \sysname evaluates the new heuristics the LLM produced, finds new adversarial samples (and maybe new regions), and suggests how to improve the heuristic. We then use the LLM to implement the suggestions and improve the heuristics in each step. The rest is similar to FunSearch (\cref{sec:background}).

We evaluate a simple execution of this design end-to-end. To mimic explanations throughout the process we ``approximate'' them: we use the adversarial inputs we found for the base heuristic as inputs to the heuristics the LLM produces in each step of the search and then map the output to the edges in the XPlain DSL (these edges capture the ``actions'' the heuristic takes~---~we ignore the node behaviors in the DSL which enforce the heuristic behavior as we have already concretized those actions). This approach is approximate because the initial adversarial regions may not contain samples that make the new heuristic underperform.

As the search progresses the heuristic improves and we have less and less samples where the heuristic underperforms. This continues until we no longer have enough information to generate meaningful suggestions~---~we no longer create suggestions after this point.

We experiment with a traffic engineering heuristic~\cite{metaopt} which ``pins'' small demands to their shortest path and optimally routes others. We use a $20$-node, $30$-link sub-graph of the CogentCo topology from the Topology Zoo~\cite{topozoo}. We impose a runtime limit of $120$ second across all heuristics, matching the runtime of the base heuristic on this topology. This implicitly gives feedback to the LLM about the runtime. 
 
 We use MetaOpt~\cite{metaopt} to find adversarial samples where the base heuristic underperforms. MetaOpt~\cite{metaopt} is a heuristic analyzer that takes a model of the optimal solution to a problem and the heuristic and finds input instances that cause that heuristic to underperform. To find multiple adversarial inputs we implement the idea in XPlain~\cite{xplain} which extends MetaOpt to find adversarial~\emph{subspaces} (regions of the input space where the heuristic underperforms).

 We use Open-AI's o4-mini and repeat each experiment $20$ times to account for the randomness in the results. To evaluate each approach we use $1500$ held-out samples (we use the same set to evaluate each approach) which we do not expose the LLM to when we create the heuristics.

 We evaluate the performance of each algorithm across $50$ ``step'' where a step is one where each island produces a new heuristic. We report ``normalized suboptimality'' values where the suboptimality is the maximum difference between the performance of the heuristic and that of the optimal algorithm, and we normalize by the suboptimality of the base seed heuristic. Our results (\autoref{table:avg-perf} and \autoref{fig:funsearch_e2e}) show~\sysname produces heuristics that have $28\times$ better worst-case performance, and $\sim200\times$ better performance on average compared those FunSearch creates. The runtime of the heuristics the two approaches create is similar.

\begin{table}[t]
\centering
\resizebox{\columnwidth}{!}{%
\begin{tabular}{lccc}
\toprule
\textbf{Method} & \shortstack{\textbf{Max suboptimality}\\(\% of base suboptimality)} & \shortstack{\textbf{Mean suboptimality}\\(\% of base suboptimality)} & \shortstack{ \textbf{Runtime}\\($\times$ the base)} \\
\midrule
\sysname & $\boldsymbol{0.5}\%$ & $\boldsymbol{0.01 \pm 0.002} \%$ & 1.00$\times$ \\
 FunSearch & 14 \%& $2.07 \pm 0.097$ \% & 0.96$\times$ \\
\bottomrule
\end{tabular}}
\caption{\sysname outperforms FunSearch. \label{table:avg-perf}}
 \vspace{-\baselineskip}
\end{table}

\section{The path forward}
\label{sec:future}
We described how we can help LLMs design better heuristics and open questions in this space (\cref{sec:intuition}). We did not discuss a number of other aspects of this problem:

\noindent \textbf{Are LLMs the right method to use?} We posed a specific question: can explanations help LLM-based search algorithms create better heuristics? But it is unclear whether LLMs are the right tool to use. We think they might be: they are flexible and provide a convenient user interface. We need more research to evaluate what the potential downsides of such solutions are and their potential impact.

\noindent \textbf{Applications in other areas in networking.} Recent work has applied heuristic analyzers to failure analysis problems~\cite{raha} where they find the set of~\emph{probable} failures that cause the network to experience the worst-case performance compared to its healthy state. There may be an opportunity to extend the approaches we discuss here and use~\sysname to help with capacity planning under failures.

\noindent \textbf{The impact on average performance and runtime.} We spent most of this paper on how to make heuristics more robust. We control the runtime of the heuristics we generate through explicit timeouts which is why all of our improved heuristics have the same runtime as the base one. Our evaluation showed the new heuristics also improved average case performance. This may not always be the case and we need more research to balance between these two objectives.

\bibliographystyle{ACM-Reference-Format}


\clearpage
\setcounter{figure}{0}
\renewcommand{\thefigure}{A.\arabic{figure}}

\definecolor{SuggestFill}{HTML}{F3E8FF}   
\tikzset{
  suggestHeaderPlain/.style={
    rounded corners=2pt,
    draw=gray!60!black,
    thick,
    fill=SuggestFill,
    font=\bfseries\Large,
    inner sep=6pt,
    text=black
  }
}

\tikzset{
  shaded/.style={
    draw,
    minimum width=1.4cm,
    minimum height=.9cm,
    rounded corners,
    fill=gray!15,
    align=center,
    semithick,
    text=black,
    drop shadow={shadow xshift=0.5mm, shadow yshift=-0.5mm, opacity=0.5}
  },
  plain/.style={
    draw,
    minimum width=2.5cm,
    minimum height=1.1cm,
    rounded corners,
    align=center,
    fill=teal!5,
    semithick,
    text=black,
    drop shadow={shadow xshift=0.5mm, shadow yshift=-0.5mm, opacity=0.5}
  }
}

\tikzset{
  pattern prompt/.style={
    rectangle split,
    rectangle split parts=2,
    rectangle split part fill={teal!30,teal!5}, 
    draw,
    thick,
    rounded corners,
    text width=0.45\textwidth,
    font=\footnotesize,
    align=left,
    inner sep=4pt,
  },
  pattern prompt title/.style={
    rectangle split part fill=gray!60,
    text=white,
    font=\bfseries\footnotesize
  }
}

\definecolor{proNavy}{RGB}{34,45,65}
\definecolor{proOffWhite}{RGB}{245,247,250}
\tikzset{
  summarybox/.style={
    rectangle split,
    rectangle split parts=2,
    rectangle split part fill={proNavy,proOffWhite}, 
    draw=proNavy!80!black,
    thick,
    text width=0.22\textwidth,
    font=\footnotesize,
    align=left,
    inner sep=8pt
  }
}
\appendix
\section{Appendix}
In this section, we provide an in-depth examination of the various components of our design.

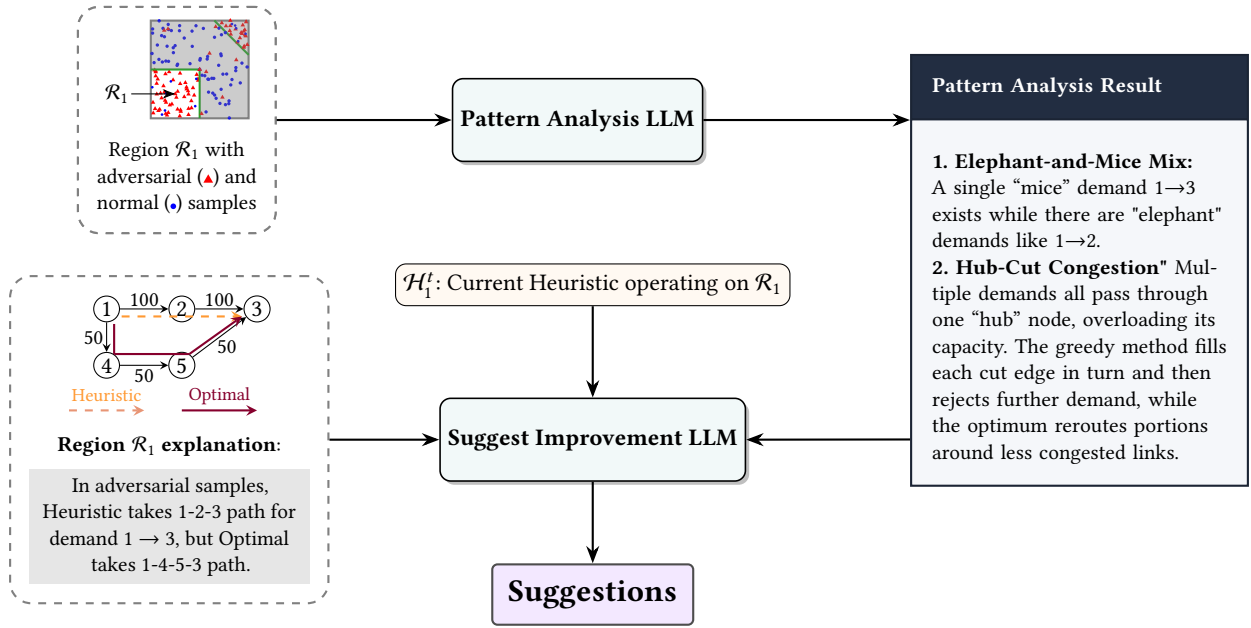
\begin{figure*}[t]
  \centering
  \small
  \begin{tikzpicture}[>=Stealth, node distance=3.2cm, scale=1]
  
    \node[inner sep=0pt]  (subspace) at (-3.5,0) { 
                    \begin{tikzpicture}[scale=0.65
                    ]
                    \draw[gray, thick] (0,0) rectangle (2,2);
                    \draw[forestgreen, thick] (1.3,2) -- (2,1.3);
                    \foreach \x/\y in {
                        0.375/0.951,
                        0.732/0.599,
                        0.156/0.156,
                        0.1/0.866,
                        0.601/0.708,
                        0.1/0.970,
                        0.832/0.212,
                        0.182/0.183,
                        0.304/0.525,
                        0.432/0.291,
                        0.612/0.139,
                        0.292/0.366,
                        0.456/0.785,
                        0.200/0.514,
                        0.592/0.046,
                        0.608/0.171,
                        0.1/0.949,
                        0.966/0.808,
                        0.305/0.098,
                        0.684/0.440,
                        0.122/0.495,
                        0.3/0.909,
                        0.259/0.663,
                        0.312/0.520,
                        0.547/0.185,
                        0.970/0.775,
                        0.939/0.895,
                        0.598/0.922,
                        0.088/0.196,
                        0.045/0.325,
                        0.389/0.271,
                        0.829/0.357,
                        0.281/0.543,
                        0.141/0.802,
                        0.075/0.987,
                        0.772/0.199,
                        0.22/0.415,
                        0.5/0.5,
                        0.55/0.55,
                        0.707/0.729,
                        0.771/0.074,
                        0.358/0.116,
                        0.863/0.623,
                        0.331/0.064,
                        0.311/0.325,
                        0.730/0.638,
                        0.887/0.472,
                        0.120/0.713,
                        0.761/0.561,
                        0.771/0.494,
                        0.523/0.428,
                        0.05/0.108
                    } {
                        \node[red, fill=red, regular polygon, regular polygon sides=3, inner sep=0.35pt, minimum size=1pt] at (\x,\y) {};
                    }
                    \foreach \x/\y in {
                         0.031/0.636,
                            0.314/0.509,
                            0.908/0.249,
                            0.410/0.756,
                            0.229/0.077,
                            0.290/0.161,
                            0.930/0.808
                    } {
                        \node[blue, fill=blue, circle, inner sep=0.35pt, minimum size=1pt] at (\x,\y) {};
                    }
                    \draw[forestgreen, thick] (0,1) -- (1,1);
                    \draw[forestgreen, thick] (1,0) -- (1,1);
                    \foreach \x/\y in {
                        1.615/1.792,
                        1.636/1.721,
                        1.849/1.755,
                        1.774/1.560,
                        1.881/1.908,
                        1.927/1.706,
                        1.734/1.826,
                        1.452/1.952,
                        1.910/1.476,
                        1.922/1.811,
                        1.626/1.894,
                        1.972/1.507,
                        1.901/1.781,
                        1.949/1.972,
                        1.822/1.645,
                        1.900/1.451,
                        1.865/1.732,
                        1.940/1.684,
                        1.626/1.999,
                        1.538/1.890,
                    } {
                        \node[red, fill=red, regular polygon, regular polygon sides=3, inner sep=0.35pt, minimum size=1pt] at (\x,\y) {};
                    }
                    \foreach \x/\y in {
                            1.725/1.899,
                            1.747/1.842,
                            1.939/1.731,
                            1.661/1.930,
                            1.788/1.577
                    } {
                        \node[blue, fill=blue, circle, inner sep=0.35pt, minimum size=1pt] at (\x,\y) {};
                    }
                    \foreach \x/\y in {
                        0.497/1.488,
                        0.067/1.140,
                        1.525/1.754,
                        0.684/1.643,
                        0.221/1.693,
                        1.595/0.300,
                        0.459/1.445,
                        1.440/1.282,
                        1.388/1.085,
                        0.363/1.817,
                        1.167/0.802,
                        0.924/1.895,
                        0.307/1.172,
                        1.012/1.223,
                        0.036/1.744,
                        1.864/1.130,
                        1.393/1.845,
                        1.414/0.305,
                        1.153/1.213,
                        0.848/1.473,
                        0.249/1.842,
                        1.740/1.038,
                        1.183/0.798,
                        1.606/0.009,
                        1.075/1.840,
                        1.475/0.904,
                        1.830/0.725,
                        1.161/1.265,
                        0.026/1.327,
                        0.356/1.922,
                        0.171/1.994,
                        1.004/1.191,
                        0.134/1.500,
                        0.420/1.796,
                        1.130/0.131,
                        1.551/0.907,
                        1.049/0.882,
                        0.802/1.119,
                        1.288/0.817,
                        1.432/1.318,
                        0.462/1.344,
                        1.600/0.357,
                        1.305/0.476,
                        1.445/1.711,
                        1.660/0.794,
                        1.336/0.410,
                        0.586/1.793,
                        0.363/1.166,
                        0.843/1.785,
                        1.635/0.684,
                        1.181/0.536,
                        1.248/0.819,
                        1.104/0.872,
                        0.589/1.897,
                        1.527/0.280,
                        1.737/0.975,
                        0.537/1.083,
                        1.267/0.516,
                        0.279/1.670,
                        1.969/1.051,
                        0.037/1.829,
                        0.236/1.153,
                        0.548/1.108,
                        1.303/1.659,
                        0.333/1.476,
                        0.166/1.206,
                        1.438/0.594,
                    } {
                         \node[blue, fill=blue, circle, inner sep=0.45pt, minimum size=1pt] at (\x,\y) {};
                    }
                    \foreach \x/\y in {
                        1.133/0.952,
                        1.327/1.874,
                        1.465/0.430,
                        0.274/1.800,
                        1.748/1.195,
                        1.201/1.330,
                        0.351/1.829,
                        1.038/0.094,
                    } {
                        \node[red, fill=red, regular polygon, regular polygon sides=3, inner sep=0.35pt, minimum size=1pt] at (\x,\y) {};
                    }
                    \node at (-0.7,0.5) {\footnotesize$\mathcal{R}_1$};
                    \draw[->, decorate] 
                     (-0.4,0.5) -- (0.5,0.5);
                  \fill[
                    black!50,       
                    opacity=0.4,    
                    even odd rule   
                  ]
                    (0,0) rectangle (2,2)
                    (0,0) rectangle (1,1)
                  ;
                    \end{tikzpicture}
                  }; 
\node[
  font=\footnotesize,
  below=5pt of subspace,
  align=center       
] (subspacenote) {%
  Region $\mathcal{R}_1$ with \\
  adversarial
  (\tikz{\node[red, fill=red,
                 regular polygon,
                 regular polygon sides=3,
                 inner sep=0.75pt] {};}) and\\
  
   normal
  (\tikz{\node[blue, fill=blue,
                 circle,
                 inner sep=0.75pt] {};}) samples%
};
\begin{pgfonlayer}{background}
  \node[
    draw=black!50,
    thick,
    dashed,
    rounded corners=8pt,
    inner sep=4pt,
    fill=white,
    fit={(subspace) (subspacenote)}  
  ] (wrapper) {};
\end{pgfonlayer}

\node[plain] (patternprompt) at ($(wrapper.east)+(4,0)$){
\textbf{Pattern Analysis LLM}};

\node[summarybox] (patternsum) at ($(patternprompt.east)+(5,-2)$) {
{\bfseries\color{white}Pattern Analysis Result}
\nodepart{two}

\textbf{1. Elephant‑and‑Mice Mix:}
   A single “mice” demand 1$\to$3 exists while there are "elephant" demands like 1$\to$2.

\textbf{2. Hub‑Cut Congestion"}
   Multiple demands all pass through one “hub” node, overloading its capacity. The greedy method fills each cut edge in turn and then rejects further demand, while the optimum reroutes portions around less congested links.
};

\node[inner sep=1pt] (dsl) at (-3.6,-3.75) { 
\begin{tikzpicture}[>=stealth, node distance=2cm]
  \node[draw, circle] (1) at (0,0.75) {1};
  \node[draw, circle] (2) at (1,0.75) {2};
  \node[draw, circle] (3) at (2,0.75) {3};
  \node[draw, circle] (4) at (0,0) {4};
  \node[draw, circle] (5) at (1,0) {5};

  \draw[->] (1) -- node[above] {\footnotesize 100} (2);
  \draw[->] (2) -- node[above] {\footnotesize 100} (3);
  \draw[->] (1) -- node[left] {\footnotesize 50} (4);
  \draw[->] (4) -- node[below] {\footnotesize 50} (5);
  \draw[->] (5) -- (3);
  \node at ($ (5) + (0.6, 0.25) $) {\footnotesize 50};
  \draw[->, orange!80, dashed, thick] ($ (1) + (0.2,-0.1) $) -- ($ (3) + (-0.2,-0.1) $);
  
  \draw[-, thick, purple!70!black] ($ (1) + (0.1,-0.2) $) -- ($ (4) + (+0.1,+0.15) $);
  \draw[-, thick, purple!70!black] ($ (4) + (+0.1,+0.15) $) -- ($ (5) + (+0.1,+0.15) $);
  \draw[->, thick, purple!70!black] ($ (5) + (+0.1,+0.15) $) -- ($ (3) + (-0.2,-0.1) $);

\draw[->, thick, set2orange, dashed] (-0.5,-0.6) -- ++(1,0); 
\node at (0,-0.4) {{\scriptsize \textcolor{orange!80}{Heuristic}}};

\draw[->, thick, purple!70!black] (1,-0.6) -- ++(1,0); 
\node at (1.5,-0.4) {{\scriptsize \textcolor{purple!70!black}{Optimal}}};
\end{tikzpicture}
};

\node[
  font=\footnotesize,
  below=5pt of dsl,
  align=center       
] (dslnote) {%
  \textbf{Region $\mathcal{R}_1$ explanation}:\\[4pt]
  \colorbox{gray!20}{%
  \centering
    \parbox{100 pt}{%
    \centering
      In adversarial samples, Heuristic takes 1‑2‑3 path for demand 1 $\to$ 3, but Optimal takes 1‑4‑5‑3 path.
    }%
  }
};

\begin{pgfonlayer}{background}
  \node[
    draw=black!50,
    thick,
    dashed,
    rounded corners=8pt,
    inner sep=4pt,
    fill=white,
    fit={(dsl) (dslnote)}  
  ] (wrapperdsl) {};
\end{pgfonlayer}

\begin{pgfonlayer}{background}
  \node[
    draw=black!50,
    thick,
    dashed,
    rounded corners=8pt,
    inner sep=4pt,
    fill=white,
    fit={(subspace) (subspacenote)}  
  ] (wrapperxplain) {};
\end{pgfonlayer}

\node[plain] (suggestprompt) at ($(wrapperdsl.east)+(3.5,0)$){\textbf{Suggest Improvement LLM}}
;

\node[
align=center,
draw=black,
fill=orange!5,
rounded corners,
inner sep=3pt
] (base) at ($(suggestprompt.north)+(0,+1.5)$) {%
$\mathcal{H}_{1}^{t}$: Current Heuristic operating on $\mathcal{R}_1$%
};



\node[suggestHeaderPlain] (suggestionsum)
  at ($(suggestprompt.south) + (0,-1.5)$) {Suggestions};
\draw[->, thick]
  (base.south) -- (suggestprompt.north);
  
\draw[->, thick]
  (wrapper.east) -- (patternprompt.west);

\draw[->, thick]
  (patternprompt.east) -- ($(patternsum.west) + (0, +2)$);

\draw[->, thick]
  ($(patternsum.west) + (0 , -2.25)$) -- (suggestprompt.east);

\draw[->, thick]
  (wrapperdsl.east) -- (suggestprompt.west);

\draw[->, thick]
  (suggestprompt.south) -- ($(suggestionsum.north)$);

\end{tikzpicture} 
  \caption{Suggester LLM Pipeline. For each region $\mathcal{R}_i$, we run a two-stage pipeline: \textbf{(1) Pattern Analysis LLM}, which, given a balanced batch of adversarial and normal samples, abstracts the failure patterns of the current heuristic; and \textbf{(2) Suggest Improvement LLM}, which, using the pattern-analysis result, Analyzer-provided explanations (\S\ref{sec:explanations}), and the code of heuristic, proposes concrete suggestions. }
\label{fig:analyzer-pipeline} 
\end{figure*}


\subsection{Suggester LLM }
\label{app:suggester}

\sysname uses a per-region Suggester LLM (\autoref{fig:architecture}) to provide suggestions for writing heuristics. 

For each region $\mathcal{R}_i$, constructed by the Heuristic Analyzer (Sec.~\ref{sec:ensemble}), we run a two–stage {\em Suggester LLM} pipeline (Fig.~\ref{fig:analyzer-pipeline}): (1) a \emph{Pattern Analysis} model that abstracts the failure patterns of the current heuristic within $\mathcal{R}_i$ (denoted by $\mathcal{H}_{i}^{t}$), and (2) a \emph{Suggest Improvement} that provides \emph{suggestions} to improve heuristic, that the Heuirstic Writer subsequently uses.

\vspace{3pt}
\noindent\textbf{Pattern Analysis LLM}. It takes as input a balanced batch of \emph{adversarial} and \emph{normal} samples in $\mathcal{R}_i$ (\autoref{fig:analyzer-pipeline}), and processes them based on the prompt given in \autoref{fig:pattern-analysis-prompt}. The prompt includes a \emph{Problem Description} that is problem-dependent and is provided by the operators.

\vspace{3pt}
\noindent\textbf{Suggest Improvement LLM}. It takes as inputs the following and processes them based on the prompt in \autoref{fig:suggest-prompt}:
\begin{enumerate}[leftmargin=1.5em, itemsep=0pt, topsep=2pt]
  \item Pattern analysis result from the pattern analysis LLM.
  \item Short, human-readable \emph{explanations} automatically produced by the Analyzer (\S\ref{sec:explanations}) (e.g., “the heuristic routes $1 \!\rightarrow\! 3$ via $1\!\rightarrow\!2\!\rightarrow\!3$, but the optimal uses $1\!\rightarrow\!4\!\rightarrow\!5\!\rightarrow\!3$ to bypass a congested hub cut”).
  \item The code of the current heuristic $\mathcal{H}_{i}^{t}$ that has generated the adversarial samples.
\end{enumerate}

\vspace{3pt}
\noindent\textbf{Why two LLMs?}
The decomposition mirrors what we observed empirically in \autoref{fig:oneshot}: a single prompt to both (i) distill adversarial patterns and (ii) give suggestions to improve the heuristic is too much to ask the LLM.
Separating the \emph{failure patterns} from the \emph{code suggestion} yields ideas that are (a) more focused, (b) targeted at the root cause of underperformance, and (c) less entangled with superficial features.

\subsection{Heuristic Writer}
\label{app:writer}

\sysname's \emph{Heuristic Writer} is a FunSearch-style loop that repeatedly asks an LLM to produce a \emph{new} heuristic from $k$ parents. In each step, it evaluates the new heuristic on a training batch (randomly selected from samples of the region), and keeps the new heuristic only if it improves the current worst-case gap on the batch or is diverse from the rest. We pass some of the worst adversarial samples from the parents and the suggestions to the LLM (Fig.~\ref{fig:mutation-prompt}). If the new heuristic fails, we attempt to fix it for a small number of times (e.g., 3 times) using a fix prompt call to an LLM (\autoref{fig:fix-prompt}).

\begin{algorithm}[H]  
\small
\caption{\sysname's Heuristic Writer}
\label{alg:writer}
\begin{algorithmic}[1]
\Require $I$ (Num islands), $T$ (max iterations)
\Require $\mathcal{H}_0$ (base heuristic to seed all islands)
\Require $\mathcal{D}_{\text{train}}$ (training batch), $\mathcal{D}_{\text{held}}$ (held-out set)
\Require $m$ (worst-$m$ samples to show the LLM)
\Require $\Pi_{\text{mut}}$ (mutation prompt template (\autoref{fig:mutation-prompt}))
\Require $\Pi_{\text{fix}}$ (fix prompt template (\autoref{fig:fix-prompt}))
\Require $\mathcal{S}$ (Suggestions generated by Suggester LLM)
\Require $\textsc{Compile}$, $\textsc{Sim}$ (compiler and simulator/evaluator)
\Require $R_{\text{fix}}$ (max automatic fix rounds)
\Require $A$ (archive size/pruning budget)
\Require $p$ (patience/early-stop window)
\Require $\textsc{Diverse}(\cdot)$ (diversity predicate of new code)
\Ensure Best heuristic ($\mathcal{H}^*$) and its held-out performance.
\State Initialize $I$ islands with the base heuristic $\mathcal{H}_0$
\For{$t \gets 1$ to $T$}
  \ForAll{islands $j \in \{1,\dots,I\}$}  \Comment{run in parallel}
    \State Select best parents in island $j$ by tournament on worst-case gap.
    \State Build mutation prompt (parent code, worst-$m$ samples, $\mathcal{S}$) using $\Pi_{\text{mut}}$.
    \State Query LLM to get candidate code; compile and simulate on $\mathcal{D}_{\text{train}}$.
    \If{candidate fails to compile/simulate}
        \State Attempt up to $R_{\text{fix}}$ automatic fix rounds.
    \EndIf
    \If{candidate improves worst-case gap (or ties but is \textsc{Diverse})}
        \State Add it to the island $j$.
    \EndIf
  \EndFor
  \State Prune archives to size $A$, re-initiate islands, and checkpoint
  \If{no improvement for $p$ iterations or worst-case gap == 0}
    \State \textbf{break}
  \EndIf
\EndFor
\State \Return best heuristic overall; report held-out performance on $\mathcal{D}_{\text{held}}$.
\end{algorithmic}
\end{algorithm}

\noindent{\emph{Runtime choice.}}
Keeping $k{=}1$ (one parent $\rightarrow$ one child) and a small number of islands keeps the search simple, cheap, and interpretable.

Fig.\autoref{fig:better_code1} shows an improved heuristic for region 1 that reduces the train and held-out gap to zero.

\subsection{\sysname's Ensemble Heuristic}
\label{app:ensemble}

Instead of using to a single global heuristic, \sysname uses a \emph{regional ensemble} heuristic $\mathcal{E}$, where 
\[
\mathcal{E} \;=\; \{(\mathcal{R}_i, \mathcal{H}_i^\star)\}_{i=1}^N,
\]
where each region $\mathcal{R}_i$ is a region of input space discovered by the Heuristic Analyzer (\S\ref{sec:ensemble}/\S\ref{app:suggester}), and $\mathcal{H}_i^\star$ is the best heuristic \emph{specialized} to that region, obtained by evolving the common base $\mathcal{H}_0$ with the Suggester + Writer loop (\S\ref{alg:writer}). For each input, \sysname runs the corresponding heuristic for the region to which the input belongs.

\paragraph{From $\mathcal{H}_0$ to $\mathcal{H}_i^\star$.}
Starting from the shared base heuristic $\mathcal{H}_0$, we run, for each region $\mathcal{R}_i$, the two–stage Suggester (pattern analysis $\rightarrow$ suggestions) followed by the Heuristic Writer (\S\ref{app:writer}). The Writer produces a sequence
$\mathcal{H}_0 \!\rightarrow\! \mathcal{H}_i^{1} \!\rightarrow\! \dots \!\rightarrow\! \mathcal{H}_i^\star$
that monotonically reduces the worst-case gap inside $\mathcal{R}_i$ on training batch.

\begin{figure*}[t]
  \centering
  \begin{tikzpicture}[scale=1]
    \node[pattern prompt, text width=0.8\textwidth] (patternprompt){
      \textbf{Pattern Analysis Prompt}
      \nodepart{two}
      
      \textbf{Problem Description}:\\[2pt]
      You are an expert in analyzing heuristic performance difference between the optimal solution and the heuristic solution in the Traffic Engineering problem. In this problem, we have a topology with nodes and directed edges with limited capacity. The inputs are the demands between the nodes. The goal is to route the maximum amount of traffic between the nodes in the network. Your final goal is to help design a better heuristic. Be concise and to the point.\\[4pt]

      \textbf{Instructions}:\\[2pt]
      Please analyze these samples and identify patterns causing performance gaps between the heuristic and the optimal solution:\\[4pt]

      \textbf{Tasks}:\\[2pt]
      1.\ Compare the adversarial and non‑adversarial sample sets (top \texttt{num\_samples} each) and list patterns that correlate with a large heuristic‑optimal gap.\\[2pt]
      2.\ For each pattern, provide a concise natural‑language description.\\[2pt]
      3.\ Combine the findings with region description (green boundary).\\[4pt]

      \texttt{Examples of adversarial
        (\tikz{\node[red, fill=red,
                       regular polygon,
                       regular polygon sides=3,
                       inner sep=0.75pt] {};}) samples: …}\\
      \texttt{Examples of normal (\tikz{\node[blue, fill=blue,
                       circle,
                       inner sep=0.75pt] {};}) samples: …}
    };
  \end{tikzpicture}
  \caption{\centering Pattern Analysis Prompt}
  \label{fig:pattern-analysis-prompt}
\end{figure*}

\begin{figure*}[t]
  \centering
  \begin{tikzpicture}[scale=1]
    \node[
      rectangle split,
      rectangle split parts=2,
      pattern prompt,
      text width=0.76\textwidth,
      inner sep=8pt
    ] (suggestprompt) at (5.8,-8) {
      \textbf{Suggest Improvement Prompt}
      \nodepart{two}
      \begin{varwidth}{\textwidth}
       \textbf{Problem Description}:\\[2pt]
        You are an expert in analyzing heuristic performance difference between the optimal solution and the heuristic solution in the Traffic Engineering problem. In this problem, we have a topology with nodes and directed edges with limited capacity. The inputs are the demands between the nodes. The goal is to route the maximum amount of traffic between the nodes in the network. Your final goal is to help design a better heuristic. Be concise and to the point.

        We have analyzed the performance of a heuristic and the optimal solution on a set of samples.\\
        \textbf{Pattern Analysis:}\\[15pt]

        \textbf{Heuristic code:}\\[15pt]

        \textbf{Explanations:}\\
        We also found out that the following decisions are the most likely to cause the gap:\\[15pt]

        \textbf{Task:}\\
        Please suggest ideas for improvements to the heuristic:

        \medskip
        \begin{tabular}{@{}ll}
          1.& What modifications could prevent these gaps?\\
          2.& What additional network metrics should be considered?\\
          3.& What alternative routing strategies might work better?\\
          4.& How can we better handle congestion and load balancing?\\
          5.& Is there a way to run optimal on a subset of the problem? For example, a subset of demands or graph?\\
          6.& Propose ideas for improvements.
        \end{tabular}

        \bigskip
        Your task is to list up to \{n\} concrete, different idea that would reduce the gap.

        In order to do your task:

        \medskip
        \begin{tabular}{@{}ll}
          1.& Examine the adversarial patterns and the decision differences.\\
          2.& From those patterns, extract up to one improvement idea likely to improve the heuristic.\\
          3.& For each idea, provide a detailed (at least 100 words), code‑agnostic explanation and reasoning.\\
        \end{tabular}

        \bigskip
        -- Requirements --

        \medskip
        \begin{tabular}{@{}ll}
          •& Do not write code, only suggest ideas.\\
          •& Do not suggest ML approaches requiring lots of training data.\\
          •& Provide a thorough explanation of each idea.\\
          •& Explain so the reader can implement it themselves.
        \end{tabular}

        \bigskip
        -- Output Format --

        \medskip
        \begin{verbatim}
[
  {
  "idea": "...",
  "reasoning": "..."
  },
  {
  "idea": "...",
  "reasoning": "..."
  }
]
        \end{verbatim}
      \end{varwidth}
    };
    \node[
align=center,
draw=black,
fill=orange!5,
rounded corners,
inner sep=3pt,
font=\footnotesize
] at ($(suggestprompt.north)+(-5.5,-5.7)$) {%
$\mathcal{H}_0$: Base Heuristic%
};

\node[
align=center,
draw=black,
fill=proNavy,
inner sep=3pt,
font=\footnotesize
] at ($(suggestprompt.north)+(-5.1,-4.5)$) {%
\bfseries\color{white}Pattern Analysis Result
};

\node[
align=center,
fill=gray!20,
inner sep=3pt,
font=\footnotesize
] at ($(suggestprompt.north)+(-3.7,-7.2)$) {%
\bfseries\color{black}Decision differences from Heuristic Analyzer
};

  \end{tikzpicture}
  \caption{Suggest Improvement Prompt}
  \label{fig:suggest-prompt}
\end{figure*}

\begin{figure*}[t]
  \centering
  \begin{tikzpicture}[scale=1]
    \node[
      rectangle split,
      rectangle split parts=2,
      pattern prompt,
      text width=0.76\textwidth,
      inner sep=8pt
    ] (suggestprompt) at (5.8,-8) {%
      \textbf{\sysname's Heuristic Writer Prompt}
      \nodepart{two}
      \begin{varwidth}{\textwidth}
        \textbf{Problem Description}:\\[2pt]
        You are an expert in analyzing heuristic performance difference between the optimal solution and the heuristic solution in the Traffic Engineering problem. In this problem, we have a topology with nodes and directed edges with limited capacity. The inputs are the demands between the nodes. The goal is to route the maximum amount of traffic between the nodes in the network. Your final goal is to help design a better heuristic. Be concise and to the point.\\[4pt]

        \textbf{Task:}\\
        Your task is to design a new heuristic different than the \textbf{Parent} heuristics.\\[4pt]

        \textbf{Parent Heuristic (1):}\\
        Here is a parent heuristic:\\
        \verb|```python| \\
        \verb|{code}| \\
        \verb|```| \\

        -- Worst Performing Samples for the Parent Heuristic (1) -- \\
        The *Parent* heuristic performed poorly on the following samples compared to the optimal solution: \\
        \texttt{Examples of adversarial
        (\tikz{\node[red, fill=red,
                       regular polygon,
                       regular polygon sides=3,
                       inner sep=0.75pt] {};}) samples for the parent (1): …}\\

        -- Suggestions to improve the Parent Heuristic (1) -- \\
        You can use the following observations/suggestions to improve the parent heuristic:\\[19 pt]

        \textbf{... Add (k) Parents}\\[4 pt]

        -- Requirements -- \\
        Based on the parent heuristics above, first analyze the pros and cons of each, and then design a new heuristic that performs better. You can use the suggestions to improve the parent heuristics if you want. \\ [2 pt]

        -- Output Format --

        \medskip
\begin{verbatim}
[
  {
  "code": "def run_heuristic(...):\n    # Your implementation here\n",
  "reasoning": "..."
  }
]
\end{verbatim}
      \end{varwidth}
    };

\node[suggestHeaderPlain, font=\footnotesize] at ($(suggestprompt.north)+(-5.8,-9.7)$) {
Suggestions
};
  \end{tikzpicture}
  \caption{\sysname's Heuristic Writer Mutation Prompt}
  \label{fig:mutation-prompt}
\end{figure*}

\begin{figure*}[t]
  \centering

  \begin{tikzpicture}[scale=1]
    \node[pattern prompt, text width=0.76\textwidth] (fixprompt) {
      \textbf{Fix Prompt}
      \nodepart{two}
      \begin{varwidth}{\textwidth}
\small
\textbf{System Instruction}:\\
You are an expert Python developer. You are given a code that is not working as expected.\\
You are given an error message. You need to fix the code.\\[6pt]

\textbf{Code to fix}:\\
\verb|```python| \\
\verb|{self.helper_code}| \\
\verb|{code}| \\
\verb|```| \\[6pt]

\textbf{Error}:\\
\verb|```| \\
\verb|{error}| \\
\verb|```| \\[6pt]

\textbf{Task}:\\
Fix the code and return \textbf{ONLY} the complete fixed code (no fences).
      \end{varwidth}
    };
  \end{tikzpicture}

  \caption{Fix prompt given to the LLM when a mutated heuristic fails to compile or simulate.}
  \label{fig:fix-prompt}
\end{figure*}

\begin{figure*}[!t]
\captionsetup{type=figure}
\caption{\sysname's improved heuristic for region 1 ($\mathcal{H}_1$)}
\begin{tcolorbox}[width=\textwidth, colback=gray!3, colframe=gray, boxrule=0.6pt, arc=2mm]
\begin{lstlisting}[style=PythonStyle, firstline=1, lastline=120]
from typing import Dict, Any, List
from collections import defaultdict

def run_heuristic(graph: Graph) -> Dict[str, Any]:
    # Phase 1: Greedy routing on top-K shortest paths.
    demands = [d for d in graph.demands if d.demand > 0]
    for d in sorted(demands, key=lambda x: x.demand, reverse=True):
        remaining = d.demand
        try:
            paths = graph.get_shortest_paths_between_pair(d.source.name, d.target.name, k=3)
        except Exception:
            paths = graph.find_all_paths_between_pair(d.source.name, d.target.name)
        for p in paths:
            cap = p.min_remaining_capacity
            if cap <= 1e-9:
                continue
            flow = min(cap, remaining)
            if flow <= 1e-9:
                continue
            graph.add_flow_to_path(p, flow)
            remaining -= flow
            if remaining <= 1e-6:
                break

    # Phase 2: Local re-optimization around high-utilization edges.
    utilization = graph.get_edge_utilization()  # {edge: utilization}
    hotspots = sorted([e for e, u in utilization.items() if u > 0.8],
                            key=lambda e: utilization[e], reverse=True)[:3]

    for edge in hotspots:
        # Build region around the hot edge
        region = {edge.source, edge.target}
        for e in graph.edges:
            if e.source in region or e.target in region:
                region.add(e.source)
                region.add(e.target)

        region_edges = [e for e in graph.edges if e.source in region and e.target in region]
        region_demands = [d for d in graph.demands if d.source in region and d.target in region]
        if not region_demands:
            continue


\end{lstlisting}
\end{tcolorbox}
\label{fig:better_code1}
\end{figure*}

\begin{figure*}[!b]
\caption{\sysname's improved heuristic for region 1 ($\mathcal{H}_1$)  (cont.)}
\begin{tcolorbox}[width=\textwidth, colback=gray!3, colframe=gray, boxrule=0.6pt, arc=2mm]
\begin{lstlisting}[style=PythonStyle, firstline=1, lastline=120]
        # Construct subgraph
        name_map = {n.name: Node(n.name) for n in region}
        sub_edges = [
            Edge(name_map[e.source.name], name_map[e.target.name], e.capacity)
            for e in region_edges
        ]
        sub_demands = [
            Demand(name_map[d.source.name], name_map[d.target.name], d.demand)
            for d in region_demands
        ]
        subg = Graph(list(name_map.values()), sub_edges, sub_demands)

        # Reset flows in region
        for e in region_edges:
            e.flow = 0.0
        graph.active_paths = {
            p: f for p, f in graph.active_paths.items()
            if all(ed not in region_edges for ed in p.edges)
        }

        # Solve exact multi-commodity flow on the subgraph
        try:
            result = find_optimal_flows(subg)
            opt_graph = result.get("graph", subg)
            for path_sub, flow in opt_graph.active_paths.items():
                if flow <= 1e-9:
                    continue
                real_edges = []
                for e_sub in path_sub.edges:
                    real = graph.get_edge(e_sub.source.name, e_sub.target.name)
                    if real is None:
                        real_edges = []
                        break
                    real_edges.append(real)
                if real_edges:
                    graph.add_flow_to_path(Path(real_edges), flow)
        except Exception:
            continue

    # Compute and return metrics
    total_met = sum(graph.active_paths.values())
    total_unmet = sum(d.demand for d in graph.demands) - total_met
    return {
        "total_met_demand": total_met,
        "total_unmet_demand": total_unmet,
        "graph": graph
    }

\end{lstlisting}
\end{tcolorbox}
\end{figure*}

\end{document}